\documentclass[final,3p,times,twocolumn]{elsarticle}

\usepackage{amssymb}
\usepackage{rotating}
\usepackage{float}
\usepackage{amsmath}
\usepackage{xcolor}
\usepackage{listings}
\usepackage{tabularx}
\usepackage{url}
\usepackage{siunitx}
\usepackage{comment}
\usepackage{placeins}
\usepackage{enumitem}
\usepackage{xurl}
\usepackage{tikz}
\usetikzlibrary{arrows.meta, positioning, shapes.geometric, calc, fit, backgrounds}
\usepackage{booktabs}
\usepackage[hidelinks]{hyperref}

\usepackage{csquotes}

\journal{Manufacturing Systems}

\begin{document}

\begin{frontmatter}

\title{KAPPS: A knowledge-based CPPS Architecture for the Circular Factory}

\author[ifl]{Etienne Hoffmann} 
\author[itm]{Jan-Felix Klein}
\author[ifl]{Sören Weindel}
\author[wbk]{Max Goebels}
\author[wbk]{Sebastian Behrendt}
\author[ais]{Daniel Hernández}
\author[ais]{Ratan Bahadur Thapa}
\author[wbk]{Jürgen Fleischer}
\author[ifl]{Kai Furmans}
\author[ais,ecs]{Steffen Staab}

\affiliation[ifl]{organization={Institute for Material Handling and Logistics (IFL), Karlsruhe Institute of Technology},
            city={Karlsruhe},
            country={Germany}}
            
\affiliation[itm]{organization={ Department of Production Engineering, KTH Royal Institute of Technology},
            postcode={Stockholm}, 
            country={Sweden}}

\affiliation[wbk]{organization={Institute of Production Science (wbk), Karlsruhe Institute of Technology},
            city={Karlsruhe},
            country={Germany}}

\affiliation[ais]{organization={Analytic Computing, Institute for Artificial Intelligence, University of Stuttgart},
            postcode={Stuttgart}, 
            country={Germany}}
 \affiliation[ecs]{organization={Electronics and Computer Science, University of Southampton},
            postcode = {Southhampton},
            country = {United Kingdom}}


\begin{abstract}

While linear manufacturing relies on homogeneous materials and predefined process sequences, circular manufacturing reintroduces used products with heterogeneous and uncertain conditions. This shift demands manufacturing systems capable of handling variable product states, dynamically reconfigurable processes, and the integration of human and machine knowledge. Conventional manufacturing IT architectures, designed for stable structures and deterministic execution, are unable to meet these requirements, as they cannot adequately represent and manage the uniqueness of individual components at runtime.

Following a design science methodology for developing a Cyber Physical Production System for circular manufacturing, we derive 14 requirements from five complementary perspectives.

Based on these requirements, we design KAPPS, a knowledge-based architecture that uses an ontology-grounded knowledge graph as a unifying data backbone, combined with a semantic interface layer to enable consistent data and information integration, reasoning, and communication across heterogeneous systems and services, turning the knowledge graph from an integration layer into the factories authoritative write-time state.

KAPPS incorporates modules for constraint enforcement and event-driven planning, enabling incremental adaptation of execution plans under uncertainty and human-machine knowledge exchange. The applicability of KAPPS is demonstrated through two implemented use cases: (i) Anomaly detection and learning through knowledge graph mediated services and (ii) runtime constraint enforcement in a modular conveyor system. Subsequently, the architecture is evaluated against the 14 requirements. The two use case implementations demonstrate that KAPPS structurally satisfies all requirements by allowing for instance-level traceability, execution-time feasibility validation, and semantic mediation between heterogeneous services. We conclude outlining plans for the continued development of the system.

\end{abstract}


\end{frontmatter}



\section{Introduction } 
\label{introduction}

\noindent Growing global material demand and the associated environmental pressures pose a fundamental challenge for modern society, demanding the efficient use of resources~\cite{ImprovingResourceEfficiency2020}. Therefore, the United Nations' Sustainable Development Goal 12 places sustainable consumption and production at the center of long-term development strategies~\cite{affairsSustainableDevelopmentGoals2025}. One promising route toward sustainable production is the circular economy, which aims to decouple resource consumption from value creation~\cite{EllenMacArthurFoundation2015}. Circular economy activities can be clustered into different, so-called R-strategies which then can be ranked by degree of value retention achieved~\cite{pottingCircularEconomyMeasuring2017a}: smarter product use (RS0~--~RS2) and extended product lifespans (RS3~--~RS7) are prioritized over material~(RS8) and energy recovery (RS9). The highest-ranked activity that is designed to be carried out at a factory scale is remanufacturing (RS6)~\cite{pottingCircularEconomyMeasuring2017a}, which is relevant for original equipment manufacturers.

\subsection{The Circular Factory}
\label{sec:motivation}
\noindent Industrial remanufacturing restores used products, subsequently referred to as \emph{cores}, to like-new condition through processes such as disassembly, cleaning, inspection, and reassembly, while preserving their original functionality and quality~\cite{nasr_remanufacturing_2006, matsumoto_trends_2016}. The concept of the Circular Factory~\cite{lanza_vision_2024}, progresses this idea by enabling the production of the latest product generation while selectively reusing suitable components recovered from collected cores.

Cores exhibit a high degree of variability in condition due to their individual life cycles. In the Circular Factory, this variability is amplified by the need to assess and match recovered components across different product generations while maintaining quality constraints. 

This cross-generational integration introduces additional heterogeneity and uncertainty that propagate throughout the production process chain:
In linear manufacturing, process sequences are fixed and the capabilities required to produce a product are known in advance. Circular production requires determining the process type (\emph{which operations should be performed on this part?}), the process sequence (\emph{in what order do these operations apply?}), and the recombination of parts (\emph{which specific parts should be combined into a product?}) at runtime. Each core thus follows an individual path through the factory, dynamically determined at factory runtime as more and more information about this part becomes available.

Covering this open and instance-specific operation set with a finite resource pool requires resources whose capabilities can be reconfigured at runtime and adapt to unforeseen situations. These resources must be coordinated by online planning systems that can manage large numbers of unique and uncertain products and processes while maintaining overall factory efficiency and maximizing value retention~\cite{fleischer_self-learning_2024}.

Despite increasing automation, skilled human workers remain essential for handling disturbances that exceed current autonomous capabilities. Manual workstations equipped with sensors enable the capture and analysis of human actions allowing implicit human knowledge to be transferred into the digital production system. This interaction supports the continuous improvement of automated resources and the gradual reduction of human intervention~\cite{zaremski_learning_2024}.

Taken together, these characteristics define a class of demands that conventional Manufacturing Execution System (MES) and Cyber-Physical Production System (CPPS) architectures were not designed to meet. As the review of the state of the art in Section \ref{sec:review} outlines in detail, existing approaches address individual facets of this problem in isolation, but none jointly provides the flexible and adaptable CPPS architecture, that a Circular Factory requires. 

\subsection{Contribution}
\label{sec:contribution}

\noindent This work addresses the technological gap between the execution-time demands of the Circular Factory and the capabilities of conventional MES and CPPS architectures. Existing approaches are largely based on assumptions of process stability, uniform product structures, and predefined execution sequences. These assumptions do not hold in Circular Factory settings, where each incoming core introduces uncertainty in its condition and admissible processing trajectory. To address this gap, we propose KAPPS (Knowledge-based Architecture for Cyber-Physical Production Systems), a four-layer CPPS architecture in which an ontology-driven knowledge graph serves as a runtime representation and constraint-enforcement mechanism for heterogeneous shop floor systems and services.
The specific contributions of this work are:
\begin{enumerate}[label=\arabic*), leftmargin=*]
\item A systematic derivation of CPPS requirements for the Circular Factory from five complementary perspectives, yielding 14 design requirements (R1~--~R14), that also define evaluation criteria for architectures supporting circular production.

\item The KAPPS architecture, comprising an Ontology Layer, a Knowledge Base Layer, a Service Layer, and an Interface Layer. In this architecture, semantic models do not merely support integration but also govern process execution, enable validation, and reflect state evolution at runtime. 
\item An open-source reference implementation of the KAPPS architecture (MIT License), together with two laboratory-scale use case scenarios that exercise it under representative Circular Factory conditions: anomaly detection and learning through knowledge graph mediated services in a robotic disassembly cell (UC1), and integration of distributed control logic across a decentralized conveyor network with knowledge-base-side enforcement of physical invariants (UC2).  
\end{enumerate}

\subsection{Research Methodology \& Outline}
\label{sec:Method}

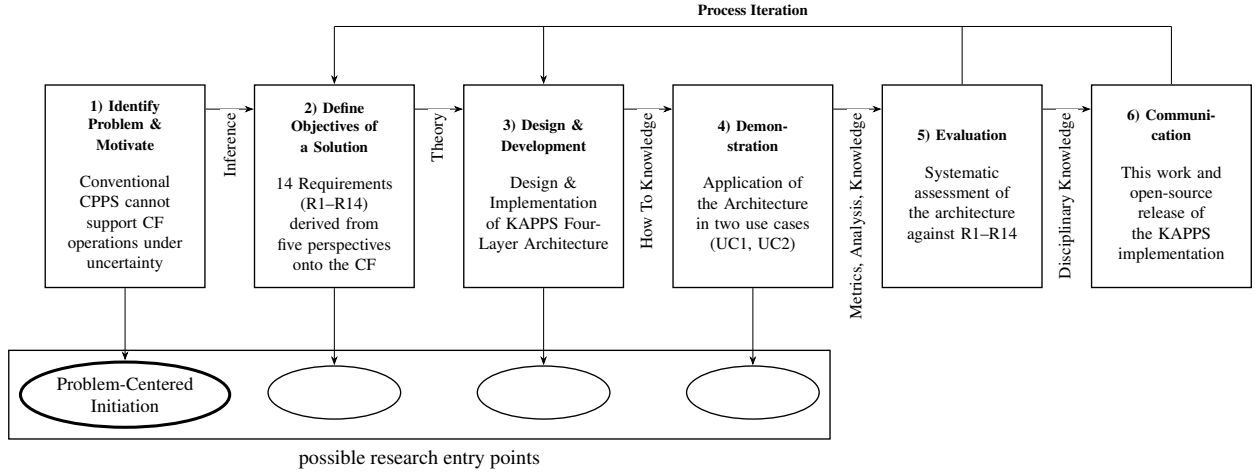
\begin{figure*}[htb]
    \centering
    
\resizebox{\textwidth}{!}{%
  \begin{tikzpicture}[
  box/.style={
    draw, thick, rectangle,
    text width=2.75cm,
    minimum height=4.2cm,
    align=center,
    inner sep=8pt
  },
  rlabel/.style={
    rotate=90,
    anchor=east,
    font=\fontsize{8}{10}\selectfont,
    fill=white,
    inner sep=2pt
  },
  ov/.style={
    draw, thick, ellipse,
    minimum width=2.75cm,
    minimum height=1.1cm,
    align=center,
    font=\fontsize{7}{9}\selectfont
  },
  arr/.style={-{Stealth[length=5.5pt, width=4pt]}, semithick},
  harr/.style={semithick},
]

\node[box] (N1) {%
  {\small\bfseries 1) Identify\\Problem \&\\Motivate}\\[10pt]
  {\fontsize{10}{10}\selectfont Conventional CPPS cannot support CF operations under uncertainty}%
};
\node[box, right=10mm of N1] (N2) {%
  {\small\bfseries 2) Define\\Objectives of\\a Solution}\\[10pt]
  {\fontsize{10}{10}\selectfont 14 Requirements (R1--R14) derived from five perspectives onto the CF}%
};
\node[box, right=10mm of N2] (N3) {%
  {\small\bfseries 3) Design \&\\Development}\\[10pt]
  {\fontsize{10}{10}\selectfont Design \& Implementation of KAPPS Four-Layer Architecture}%
};
\node[box, right=10mm of N3] (N4) {%
  {\small\bfseries 4) Demon-\\stration}\\[10pt]
  {\fontsize{10}{10}\selectfont Application of the Architecture in two use cases (UC1, UC2)}%
};
\node[box, right=10mm of N4] (N5) {%
  {\small\bfseries 5) Evaluation}\\[10pt]
  {\fontsize{10}{10}\selectfont Systematic assessment of the architecture against R1--R14}%
};
\node[box, right=10mm of N5] (N6) {%
  {\small\bfseries 6) Communi-\\cation}\\[10pt]
  {\fontsize{10}{10}\selectfont This work and open-source release of the KAPPS implementation}%
};

\draw[arr] ([yshift=1.6cm]N1.east) -- ([yshift=1.6cm]N2.west);
\draw[arr] ([yshift=1.6cm]N2.east) -- ([yshift=1.6cm]N3.west);
\draw[arr] ([yshift=1.6cm]N3.east) -- ([yshift=1.6cm]N4.west);
\draw[arr] ([yshift=1.6cm]N4.east) -- ([yshift=1.6cm]N5.west);
\draw[arr] ([yshift=1.6cm]N5.east) -- ([yshift=1.6cm]N6.west);

\node[rlabel, font=\fontsize{10}{13}\selectfont] at ($(N1.east)!.5!(N2.west)+(0,1.6cm)$) {Inference};
\node[rlabel, font=\fontsize{10}{13}\selectfont] at ($(N2.east)!.5!(N3.west)+(0,1.6cm)$) {Theory};
\node[rlabel, font=\fontsize{10}{13}\selectfont] at ($(N3.east)!.5!(N4.west)+(0,1.6cm)$) {How To Knowledge};
\node[rlabel, font=\fontsize{10}{13}\selectfont] at ($(N4.east)!.5!(N5.west)+(0,1.6cm)$) {Metrics, Analysis, Knowledge};
\node[rlabel, font=\fontsize{10}{13}\selectfont] at ($(N5.east)!.5!(N6.west)+(0,1.6cm)$) {Disciplinary Knowledge};

\coordinate (PIt2) at ([yshift=12mm]N2.north);
\coordinate (PIt3) at ([yshift=12mm]N3.north);
\coordinate (PIt5) at ([yshift=12mm]N5.north);
\coordinate (PIt6) at ([yshift=12mm]N6.north);

\draw[harr] (PIt2) -- (PIt6);
\draw[harr] (N5.north) -- (PIt5);
\draw[harr] (N6.north) -- (PIt6);
\draw[arr] (PIt2) -- (N2.north);
\draw[arr] (PIt3) -- (N3.north);
\node[above, font=\small\bfseries] at ($(PIt2)!.5!(PIt6)+(0,1mm)$) {Process Iteration};

\node[ov, below=15mm of N1, font=\fontsize{11}{13}\selectfont , line width = 1.8 pt] (O1) {Problem-Centered\\Initiation};
\node[ov, below=16mm of N2] (O2) {};
\node[ov, below=16mm of N3] (O3) {};
\node[ov, below=16mm of N4] (O4) {};

\draw[arr] (N1.south) -- (O1.north);
\draw[arr] (N2.south) -- (O2.north);
\draw[arr] (N3.south) -- (O3.north);
\draw[arr] (N4.south) -- (O4.north);

\begin{pgfonlayer}{background}
  \node[draw, thick, rectangle,
        fit=(O1)(O2)(O3)(O4),
        inner sep=6pt] (ovalbox) {};
\end{pgfonlayer}
\node[below=1mm of ovalbox, font=\large] {possible research entry points};

\end{tikzpicture}}
    \caption{The DSRM Process adopted from~\cite{peffersDesignScienceResearch2007} for the development of the KAPPS Architecture.}
    \label{fig:DSRM}
\end{figure*}

\noindent This work is conducted as a design-science effort: its objective is to construct and evaluate an artifact (a knowledge-based CPPS architecture)~\cite{hevnerDesignScienceInformation2004}. To structure both the research process and its presentation, we adopt the Design Science Research Methodology (DSRM) of Peffers et al.~\cite{peffersDesignScienceResearch2007}, which organizes design-science work into six sequential activities: 1) problem identification and motivation, 2) definition of the objectives of a solution, 3) design and development, 4) demonstration, 5) evaluation and 6) communication, with four admissible entry points and an explicit feedback loop from evaluation back to design and development. We refer to Peffers et al.~\cite{peffersDesignScienceResearch2007} for a full discussion of the methodology and its rationale.
The present work is a problem-centered initiation in the sense of the DSRM: it begins from the observation introduced in Section \ref{sec:motivation} and proceeds through the nominal DSRM sequence.
The version presented in this paper is the second iteration of the conception of the KAPPS architecture. A first iteration was performed internally before, and its evaluation produced findings that shaped the present design. 

The remainder of this Paper is structured according to the DSRM Process as  shown in Figure~\ref{fig:DSRM}:
\begin{itemize}
    \item Section~\ref{sec:perspectives} addresses Activity 1) and 2) by analyzing the Circular Factory~\cite{lanza_vision_2024} through five complementary perspectives that jointly define the requirements imposed on a supporting CPPS. The requirements are derived from the body of publications describing the Circular Factory according to~\cite{lanza_vision_2024} and  summarized in Table~\ref{tab:requirements}, closing the Section.
    \item Section~\ref{sec:review} completes Activity 1) by reviewing the state of the art in CPPS architectures and synthesizing a gap analysis against the derived requirements, thereby establishing which of the derived objectives are not jointly addressed by existing paradigms.
    \item Section~\ref{sec:architecture} presents Activity 3) in the form of the four-layer KAPPS architecture. 
    \item Section~\ref{sec:demonstration} instantiates Activity 4) through demonstration of two use cases.
    \item Section \ref{sec:evaluation} presents Activity 5): evaluation is conducted as a systematic comparison of the observed results from Section \ref{sec:architecture} \& \ref{sec:demonstration} against the objectives R1–R14 defined in Section \ref{sec:perspectives}.
    \item Section \ref{sec:limitations} delimits the scope within which the present evaluation holds and records the feedback from Activity 5) into the design of a subsequent iteration.
    \item Section \ref{sec:conclusion} concludes the paper and illustrates plans for future work.
\end{itemize}
\FloatBarrier
\section{The Circular Factory and its requirements on a CPPS}
\label{sec:perspectives}

\noindent 

To structure the analysis of the Circular Factory, this section adopts five complementary perspectives. Each perspective isolates a specific dimension of the problem space and serves as a basis for deriving the capabilities required of a CPPS.

\subsection{Perception Perspective (P1)}
\label{sec:perception}
\noindent In linear production, perception serves as a conformance check: inspection points are coupled to predefined process steps and verify that each operation meets specification~\cite{grooverAutomationProductionSystems2019}. The Circular Factory shifts this role. Process sequences are determined during operation, and perception provides the information necessary to decide subsequent steps. 
Inspection plans are therefore not fixed in advance but generated at runtime based on the current information state of each instance, enabling the selective acquisition of additional evidence where uncertainty remains high~\cite{kaiser_semantic_2026}.

Evidence sources differ both in what they observe and in how observations are represented. The information state of an instance must abstract from this heterogeneity by capturing what is currently known, independent of the originating source and the representation format~\cite{lanza_vision_2024,heizmannManagingUncertaintyProduct2024} \textbf{(R1)}.

Perception is a continuous process in which heterogeneous evidence from sensing, process monitoring, and evolving resource states is incrementally fused. This fusion must be probabilistic to preserve the uncertainty inherent in each evidence stream ~\cite{heizmannManagingUncertaintyProduct2024}, while the resulting state must remain coherently structured and queryable representation so that planning and execution operate on a consistent view of the instance \textbf{(R2)}.

While iterative measurements can reduce uncertainty, each measurement also contributes its own uncertainty which must be explicitly propagated~\cite{hoffmannEfficientUncertaintyPropagation2026}. Consequently, uncertainty must be represented at the attribute-level, collectively defining the current information state of the instance~\cite{heizmannManagingUncertaintyProduct2024} \textbf{(R3)}.

Together, R1-R3 establish the instance information state as a central element of the CPPS: a continuously updated system-level representation that is consistently available to downstream planning and execution.

\subsection{Product Perspective (P2)}
\label{sec:product}
\noindent In the Circular Factory, products are composed of both recovered and newly manufactured components with varying condition and origin. As a result, the functional behavior and quality of the final product depend not only on the condition of individual components but also on their compatibility and collective performance. To manage these dependencies, product knowledge must be organized along two complementary levels: an \emph{instance level}, capturing the current state of each individual component, subsystem, or product; and a \emph{type level}, capturing the generic knowledge applicable to all products of a given kind~\cite{graubergerEnablingVisionPerpetual2024}. 
To realize the instance-level view, every component must be uniquely identifiable and continuously traceable from initial inspection and disassembly through rework, reassembly, and integration into a new product~\cite{graubergerEnablingVisionPerpetual2024}. This requires a persistent digital representation for each physical component, with a unique identifier and a bidirectional link between the physical entity and its digital counterpart~\textbf{(R4)}.

The information state of an instance evolves over time, as new evidence becomes available. Ensuring consistent integration requires that each contribution is recorded together with its provenance, specifying which observation produced which update ~\cite{heizmannManagingUncertaintyProduct2024}. The CPPS must therefore link each instance representation not only to its current information state but also to the underlying observations and their provenance. This enables the construction of the information state to be inspected, audited, and revised when new evidence supersedes earlier assumptions~\textbf{(R5)}.

A type-level view is required to represent generic product knowledge, including reference architectures, functional dependencies, and design knowledge~\cite{hemmerichHowDealProducts2025}, as well as tolerance schemes that constrain the fit of components and cross-generational compatibility constraints~\cite{graubergerEnablingVisionPerpetual2024}. To make this knowledge operational, the CPPS must provide explicit links between instance-level representations and type-level product models, which define the structural, functional, and quality constraints applicable to all instances of a given kind~\textbf{(R6)}.

\subsection{Production Planning, Control and Execution Perspective (P3)}
\noindent
As they pass through the factory, sourced cores undergo sequences of transformative operations that modify their geometry and functional properties, such as repairing damaged features or removing components~\cite{fleischer_self-learning_2024}. Since these processes directly affect the feasibility of subsequent operations, they must follow a well-defined sequence: as an example, a disassembly operation that requires a component to be present cannot be performed once that component has already been removed.

From a planning perspective, the central task is to coordinate production activities across both linear and circular flows~\cite{lanza_vision_2024}. Planning decisions must be grounded in the current information state: each operation is only admissible if the required process knowledge is available, and the required resource provides the necessary capabilities~\cite{bailControlArchitectureRobust2025, pfrommerOntologyRemanufacturingSystems2022}. Therefore, the CPPS must ensure that all planned operations are feasible with respect to the current information state before they are committed to execution~\textbf{(R7)}.

However, as the information state evolves continuously, as described in Section \ref{sec:perception}, static plans are insufficient. Instead, execution must be adaptive: state changes or the completion of  operations must directly subsequent actions and decision processes. The CPPS requires support of event-driven execution, enabling state changes and operation completions to trigger subsequent actions, notifications, or decision processes~\textbf{(R8)}.

Even with feasible plans and adaptive execution, correctness is not guaranteed. Planning operates on a static information state, whereas execution enacts physical transformations that must obey real-world constraints. For example, components do not self-duplicate, do not teleport between non-adjacent locations, and cannot undergo operations that presuppose components or resource states that are no longer present~\cite{bailControlArchitectureRobust2025}. The CPPS must therefore enforce physical feasibility constraints on all state transitions at the moment of their execution, independently of the planning logic that produced them~\textbf{(R9)}.

\subsection{Resource Perspective (P4)}

\noindent The Circular Factory relies on heterogeneous production, inspection, and intralogistics resources that collectively provide the capabilities required to realize its operation~\cite{lanza_vision_2024}. To cope with the previously discussed uncertainties, these resources must be highly adaptive and reconfigurable~\cite{fleischer_self-learning_2024, lanza_vision_2024}.
Achieving such adaptability requires resources to expose their capabilities through service-based representation. These services  may be deployed across different system levels, ranging from edge controllers embedded in individual resources to factory-level orchestration services~\cite{bailControlArchitectureRobust2025}. Because these services operate concurrently on a shared system state, consistency between local service views and the global factory state becomes critical. In distributed systems, it is well understood that strong consistency, system-wide coordination, and low-latency updates cannot be guaranteed simultaneously under all conditions \cite{abadi_consistency_2012}. Without consistent state alignment, services may trigger actions that are valid under outdated local views but invalid with respect to the current global state. The CPPS must therefore ensure coordinated execution of distributed services based on a consistent view of the information state \textbf{(R10)}.

The need for persistent identity and traceability applies not only to product instances but also to the resources that operate on them. Resources must therefore be represented as persistent, uniquely identifiable instances. Only then can execution services reliably determine which resource provides which capability, which resource is currently engaged with which product, and which resource is responsible for a given operation~\cite{pfrommerOntologyRemanufacturingSystems2022,bailControlArchitectureRobust2025,kleinKnowledgeBasedIntralogisticSystem2025}.
R4 must therefore be generalized to include all relevant assets on the shop floor. 
Finally, shop-floor resources communicate via heterogeneous industrial protocols and data formats~\cite{bailControlArchitectureRobust2025,kleinKnowledgeBasedIntralogisticSystem2025}.
To integrate these into a unified system view, the CPPS must provide mediation between heterogeneous communication protocols and the system's information state~\textbf{(R11)}.

\subsection{Learning Perspective (P5)}

\noindent The Circular Factory is a learning system by design, in which operational experiences is continuously transformed into improved production capabilities~\cite{lanza_vision_2024}. This includes extracting knowledge from human behavior with the goal of progressively transferring it into autonomous system capabilities~\cite{zaremski_learning_2024, fleischer_self-learning_2024}. During operation, the system generates large volumes of heterogeneous data, including execution traces, sensor observations, and interaction patterns~\cite{zaremski_learning_2024, heizmannManagingUncertaintyProduct2024}. These data form the foundation for experience-driven learning that enhances autonomy, robustness, and adaptability over time~\cite{fleischer_self-learning_2024, koch_enhancing_2025, klein_toward_2026}.

For this learning process to be effective, factory operations must be inferrable from its information state~\cite{bailControlArchitectureRobust2025,pfrommerOntologyRemanufacturingSystems2022}. In particular, the decisions that transform a product instance from one state into another constitute a primary learning signal for system evolution~\cite{zaremski_learning_2024}. The CPPS must therefore preserve decision-level provenance, capturing each decision together with its actor, context, observed state, and outcome~\textbf{(R12)}.

Beyond raw traces and decisions, learning in the Circular Factory must produce reusable abstractions that generalize experience. As an example, enabling the transfer of human-performed tasks to automated execution requires deriving task structures and temporal constraints from observed interactions with the products~\cite{dreherLearningSymbolicSubsymbolic2024, zaremski_learning_2024}. These models must be integrated into the system’s information state and made directly available to factory-level planning and execution. The CPPS must therefore treat learned abstractions as first-class elements of its information state, on equal footing with the products, processes, and resources knowledge~\textbf{(R13)}.

To support learning over long system lifetimes, the Circular Factory must account for continuous evolution of products, resources, and learned models~\cite{vogel-heuser_long_2024,lanza_vision_2024}. In such settings, maintaining only the current system state is insufficient, as leaning requires understanding how past conditions, actions, and decisions led to observed outcomes. The CPPS must therefore be able to reconstruct past configurations of the factory, enabling retrospective analysis and the re-evaluation or retraining of models against historical ground truth ~\textbf{(R14)}.

\begin{table*}[ht]
    \centering
    \begin{tabularx}{\textwidth}{p{1.5cm}| p{1.5cm} |X} \hline
        \textbf{Req.} & \textbf{Persp.} \\ \hline
        R1 & P1  & Information-centric representation of heterogeneous data. 
        \\ \hline
        R2 & P1 & Unified and centrally queryable information state aggregated from heterogeneous data sources.         
        \\ \hline
        R3 & P1    &  Flexible attribute-level uncertainty representation with a queryable best estimate and incremental refinement. 
        \\ \hline
        R4 & P2, P4 & Persistent, uniquely traceable digital representation of all assets on the shopfloor. 
        \\ \hline
        R5 & P2 & Linkage of each instance representation to its underlying observations and to the provenance of their integration into the information state.
        \\ \hline
        R6& P2 & Referable type-level product model integration.
        \\ \hline
        R7 & P3 & Feasibility validation of planned operations against the current information state.
        \\ \hline
        R8 & P3 & Event-driven execution based on state changes.
        \\ \hline
        R9 & P3 & Execution-time enforcement of physically feasible state transitions. 
        \\ \hline
        R10 & P4 & Coordination of distributed execution services over a consistent shared view of the information state. 
        \\ \hline
        R11 & P4 & Mediation between heterogeneous industrial communication protocols and the system's information state. 
        \\ \hline
        R12 & P5 & Preservation of decisions with their actor, context, observed state, and outcome.  
        \\ \hline
        R13 & P5 & Persistence and reuse of learned abstractions as first-class entries in the information state.
        \\ \hline
        R14 & P5 & Temporal reproducibility of past system states for retrospective analysis and learning.
        \\ \hline
    \end{tabularx}
    \caption{Summary of requirements on a CPPS, derived from the multiple perspectives on the Circular Factory}
    \label{tab:requirements}
\end{table*}

\subsection{Conclusion}
\noindent
The five perspectives jointly yield the 14 requirements summarized in Table~\ref{tab:requirements}.
Rather than forming an independent list of functional features, these requirements jointly constrain the CPPS architecture: no subset is sufficient in isolation. Representational requirements without execution-time enforcement allow physically infeasible states; learned abstractions without temporal reproducibility become detached from the conditions under which they were generated; and event-driven execution without coordination over a consistent system enables distributed services to act on divergent local states. It is this interdependence that necessitates the integrated architecture presented in the remainder of the paper.
The present work does not introduce new perception or inference mechanisms, planning algorithms, production resources, communication protocols, execution logic, or learning algorithms. Instead it argues that scalable operation in the Circular Factory requires a CPPS that satisfies R1-R14 as structural properties of the system architecture. The derived requirements serve as the evaluation criteria in Activity 5) of the DSRM, see Section \ref{sec:Method}.
Whether and to what extent existing CPPS architectures satisfy these 14 requirements is addressed in the following section, which reviews existing architectural paradigms regarding their suitability to fulfill the requirements of the Circular factory and discusses the role of knowledge graph extensions to them.

\FloatBarrier

\section{State of the Art in CPPS Architectures}
\label{sec:review}

\noindent CPPS architectures differ in what they treat as the substrate of runtime authority, conceived as the part of the system that is written to, synchronized against, or consulted when an execution-time decision is committed.

Four paradigms organize the field along this axis: control-centric architectures locate authority in distributed decision entities; data-centric architectures locate it in standardized digital representations; model-centric architectures locate it in engineering models and digital-twin-centric architectures locate it in per-asset virtual replicas.knowledge based extensions can be combined with every one. RAMI~4.0~\cite{DINSPEC91345} sits outside this taxonomy: it provides a reference framework on which these paradigms are positioned~\cite{yli-ojanperaAdaptingAgileManufacturing2019}, not a substrate of its own. The taxonomy is analytically distinct, though the AAS-RDF hybrids examined in Section~\ref{sec:sota-knowledge} cross its boundaries; the crossing bears directly on what each paradigm structurally cannot provide for the requirements derived in Section~\ref{sec:perspectives}.

\subsection{Control-centric architectures}
\label{sec:sota-control}
\noindent
Control-centric architectures derive system behavior from the interaction of distributed decision entities. Holonic manufacturing~\cite{vanbrusselReferenceArchitectureHolonic1998} established the paradigm's foundational realization through three cooperating holon types (order, product, and resource), each combining \emph{autonomy} with \emph{co-operation} and exchanging process knowledge, production knowledge, and process-execution knowledge to drive production. Cooperation among holons is governed by basic rules that bound their individual autonomy~\cite{vanbrusselReferenceArchitectureHolonic1998}, but those rules operate at the decision layer of each agent rather than at a shared substrate. Recent CPPS realizations refine this pattern along two lines. The holonic line continues to elaborate the original commitment: Najjari et al.~\cite{najjariCyberphysicalProductionSystems2021a} introduce hybrid compositions that combine hierarchical coordination at the management and supervision layers with peer-to-peer execution agents; Macherki et al.~\cite{macherkiQHARQHolonicBasedARchitecture2021} extend the holon concept across physical, cyber, human, and energy dimensions through holonic self-configuration. The broader multi-agent-systems line develops the paradigm with explicit attention to coordination protocols and supporting knowledge representation: Müller et al.~\cite{mullerArchitectureKnowledgeModelling2023} pair a multi-agent system with the MAPE-K loop for self-organized reconfiguration management, with a UML/XML information model serving as the shared knowledge substrate among the management agents and a separate service-oriented control layer below it.

Control-centric architectures are robust under local disturbance and adaptive to local reconfiguration: each agent decides on its own logic without depending on the availability or consistency of a central coordinator, so failures and changes are absorbed locally rather than propagating into a global re-orchestration. The paradigm's contribution to a CPPS architecture is this local-autonomy property, and it remains the standard against which distributed coordination is measured. The same commitment bounds what the paradigm offers as a complete architecture: with no central queryable state, there is no substrate from which the Circular Factory as a whole can be observed, no integration point for observations from heterogeneous sources, and no shared view over which distributed services could coordinate beyond bilateral negotiation. Requirements R2, R5, and R10 fall outside this paradigm's natural scope, and the lifecycle requirements (R12, R13, R14) inherit the same limit, with decisions, learned abstractions, and past system states bound to the agents that produced them.

\subsection{Data-centric architectures}
\label{sec:sota-data}
\noindent
Data-centric architectures organize the system around a standardized digital representation from which downstream services read and to which they write. The Asset Administration Shell (AAS), defined in RAMI 4.0~\cite{DINSPEC91345} as the Industry 4.0 Component standard, has emerged as the dominant realization of this substrate~\cite{neubauerArchitectureManufacturingXBringing2023}. AAS alone is not an architecture; orchestration is provided by frameworks such as the FA3ST service~\cite{jacobyFA3STServiceOpen2022}, which implements the reactive AAS as an open-architecture component exposing standardized REST and OPC UA APIs, and aas-middleware~\cite{behrendtAasmiddlewareEnablingInteroperable2025}, which structures service-oriented integration through Connector, Datamodel, Formatter, Mapper, and Workflow abstractions.
AAS-based architectures excel on interoperability across the IT-OT divide: the AAS is a federated standard substrate that absorbs protocol heterogeneity at the boundary through connector frameworks and integrates with existing industrial tooling through established exchange formats~\cite{rongenModellingAASRDF2023}. Their limits follow from the substrate's shape. An AAS is a per-asset tree built from predefined submodel templates. Information shared across assets is not native to the substrate: the same attribute carried by two assets is duplicated in each shell or routed through a Reference Element whose de-referencing remains poorly supported by current tooling~\cite{rongenModellingAASRDF2023}. Cross-asset relations have no expression at the substrate level, because each shell is a rooted tree rather than a node in a shared graph. Writes are admitted on per-submodel structural validity rather than on cross-entity feasibility.
Observations and their provenance, decisions with their actor, context, and outcome, and learned abstractions extracted across multiple assets have no native representation in the federated substrate, and the unified centrally queryable state required by R2 cannot be assembled from a collection of rooted trees without external aggregation.

\subsection{Model-centric architectures}
\label{sec:sota-model}
\noindent
Model-centric architectures locate runtime authority in engineering models. It is the longest-established paradigm in industrial automation: the enterprise-control hierarchy on which industrial automation has been built for three decades, standardized as IEC 62264~\cite{IEC62264-1}, is model-driven by construction, and the four other paradigms can be read as departures from this layered structure. Within CPPS, the canonical realization is standardized as IEC 61499: distributed control through event-driven function blocks, in which the function block diagram is both the engineering model and the runtime~\cite{IEC61499-1}. Pérez et al.~\cite{perezCPPSArchitectureApproach2015} introduced this commitment as a CPPS architecture for Industry 4.0, da Rocha et al.~\cite{darochaIntegratingIEEE14512022} develop the most integrated realization to date by composing IEC 61499 with smart-sensor descriptions inside an industrial reference framework, and Torres et al.~\cite{torresLiteratureReviewCloudBased2025} survey the cloud-based service-oriented variants and confirm that this remains the dominant model-centric tradition in CPPS.

While model-centric approaches are still broadly adopted in industry due to decades of standardization tooling, deterministic execution semantics, and design-time verifiability, they are inflexible by design: the engineering model is authored at design time, and operational reality that the model did not anticipate cannot enter the substrate without re-engineering. Observations and their provenance have no representation alongside the model's typed structure; decisions made by services or operators, with their actor, context, and outcome, have no representation either; and learned abstractions extracted from operational data cannot enter the substrate as first-class entries without being re-engineered as new model elements. The model is an artifact of engineering, not a substrate of operations.

\subsection{Digital-twin-centric architectures}
\label{sec:sota-dt}
\noindent
Digital-twin-centric architectures locate runtime authority in virtual replicas of physical assets. Grieves~\cite{grieves2014digital} introduced the foundational three-part construct of physical asset, virtual asset, and the data flow that links them; Kritzinger et al.~\cite{kritzinger2018digital} provide the definition the field has converged on, in which a digital twin requires bidirectional automated flow with the digital object able to act as a controlling instance over its physical counterpart. Talkhestani et al.~\cite{talkhestaniArchitectureIntelligentDigital2019a} make the structural commitment of the paradigm explicit at CPPS scope: a system-level twin is constructed as a composite of individual digital twins that exchange data and participate in interdisciplinary co-simulation. Park et al.~\cite{parkDigitalTwinbasedCyber2020} provide an integrated realization in which their Product, Process, Plan, Plant, and Resource (P4R) information model separates type-level classes from instance-level state, serving as the closest analogue in the digital-twin literature to the PPR formalism this paper builds on.

While digital-twin-centric architectures support real-time control authority over individual physical assets through Kritzinger's bidirectional synchronization criterion, the structural commitment that defines the paradigm also limits it. A digital twin is scoped to its physical referent. Cross-asset relations exist as exchanged data and co-simulated interactions, not as shared state with shared admission, so a write that affects a workpiece, the machine processing it, and the fixture holding it has no twin that owns the joint configuration. The paradigm cannot deliver the consistent shared view across distributed services required by R10, and cannot enforce the cross-PPR-instance constraints of R7. The same engineering-artifact rigidity that constrains model-centric architectures recurs here: the model a digital twin runs on is authored at engineering time, and runtime synchronization fills in instance state but cannot evolve the type-level structure.

\subsection{Knowledge-based Extensions}
\label{sec:sota-knowledge}
\noindent

The four paradigms surveyed in Sections~\ref{sec:sota-control}–\ref{sec:sota-dt} share a structural deficit: none provides a substrate in which inter-asset and cross-entity relationships are first-class. The literature's response has been to extend architectures of every paradigm with a knowledge graph as an informative layer. Da Silva et al.~\cite{dasilvaMappingCapabilitySkill2023} develop bidirectional declarative mappings between AAS submodels and a capability-and-skill ontology. Westermann et al.~\cite{westermann_representing_2023} integrate learned timed automata with a formal knowledge graph composed of AutomationML, SOSA/SSN, UML state machine, and ISO 17359 ontologies. Müller et al.~\cite{mullerContextenrichedModelingUsing2022} construct a context-enriched knowledge graph that augments an intelligent digital twin with graph embeddings. Across paradigms, the recurring move is the same: keep the paradigm's primary substrate, add a knowledge layer alongside it.

This layer is not idle. Wan et al.~\cite{wanContextawareSchedulingControl2022} demonstrate that a rules-based reasoner over a manufacturing ontology emits events that invoke self-adaptation on a cognitive controller or trigger rescheduling on a context-aware scheduler. Novák et al.~\cite{novakMitigatingUndesiredConditions} demonstrate that a Product–Process–Resource Asset Knowledge Graph supports two operator-facing capabilities: capability matchmaking for dynamic resource allocation, and LLM-mediated root-cause analysis of undesired conditions. Järvenpää et al.~\cite{jarvenpaaSemanticRulesCapability2023} demonstrate that SPIN rules over OWL ontologies infer combined capabilities of resource compositions for capability matchmaking during system design.

The presented related work shows, that knowledge graphs in CPPS already deliver reasoning-driven control invocation, diagnostic support, and capability-based resource allocation, going beyond passive representation and interlinking. It therefore already fulfills several of the requirements derived in Section~\ref{sec:perspectives} individually; the knowledge graph as a state representation fulfills many more, as Section~\ref{sec:evaluation} demonstrates. The last column of Table~\ref{tab:paradigm-coverage} previews this reach: the structural capabilities of the knowledge graph itself, enriched by what the related work presented evidence beyond it.

\begin{table}[ht]
\centering
\centering
\caption{Structural reach of CPPS architecture paradigms across requirements R1--R14. Columns 1--4 show what each paradigm fulfills as a non-hybrid architecture, through its reference architecture or recent cited work, including via standard tooling. Column 5 shows the structural capabilities of the knowledge graph as state representation, enriched by what the surveyed extensions evidence beyond it individually.}
\label{tab:paradigm-coverage}
\setlength{\tabcolsep}{4.5pt}
\begin{tabular}{lcccc|c}
\toprule
 & \shortstack{Control-\\centric}
 & \shortstack{Data-\\centric}
 & \shortstack{Model-\\centric}
 & \shortstack{DT-\\centric}
 & \shortstack{KG ex-\\tensions} \\
\midrule
R1  & \checkmark & \checkmark & \checkmark  & \checkmark & \checkmark \\
R2  &            &            & \checkmark  &            & \checkmark \\
R3  &            &            &             &            & \checkmark \\
R4  & \checkmark & \checkmark & \checkmark  & \checkmark & \checkmark \\
R5  & \checkmark & \checkmark & \checkmark  &            & \checkmark \\
R6  &            & \checkmark & \checkmark  &            & \checkmark \\
R7  &            &            &             &            &            \\
R8  & \checkmark & \checkmark & \checkmark  & \checkmark & \checkmark \\
R9  &            &            &             &            &            \\
R10 &            & \checkmark &             &            & \checkmark \\
R11 & \checkmark & \checkmark & \checkmark  & \checkmark &            \\
R12 &            &            & \checkmark  &            & \checkmark \\
R13 &            &            &             &            & \checkmark \\
R14 & \checkmark & \checkmark & \checkmark  & \checkmark & \checkmark \\
\bottomrule
\end{tabular}
\end{table}
\subsection{Research Gap}
\label{sec:sota-gap}
\noindent

The extensions surveyed in Section~\ref{sec:sota-knowledge} address every requirement individually. They also reveal the structural limit: in every reviewed application, the knowledge graph is an extension, never the substrate. Decisions appear in the graph after they are made elsewhere, because the graph by itself does not provide what a CPPS architecture must provide. Distributed services need orchestration. Industrial protocols need connector-based mediation. Runtime objects need type-safe access to live operational state. These integration mechanisms have been developed by the data-centric and control-centric paradigms over decades of standardization and tooling, and they are absent from the semantic-web technology stack. The gap is therefore not technological. SHACL handles declarative constraint validation, AAS-based middleware provides service integration and protocol mediation, PPR ontologies and capability models provide the domain vocabulary, and object-graph mapping precedents demonstrate typed runtime access. What is missing is an architectural commitment that no surveyed approach has made: the knowledge graph is the write-authoritative state of the CPPS, and admission of every state change is gated by validation against its formal constraint layer. This commitment elevates the semantic layer from analytical role to runtime substrate, turns SHACL constraints into admission conditions that govern execution, and provides the single validated path through which the operational, lifecycle, and learning requirements of the Circular Factory can be jointly met. The KAPPS architecture, presented in the following section, is organized around this commitment, drawing on the integration patterns the other paradigms have developed and binding them around the knowledge graph.

\FloatBarrier
\section{The KAPPS Architecture}
\label{sec:architecture}
\begin{figure*}[ht]
    \centering
    \includegraphics[width=\linewidth]{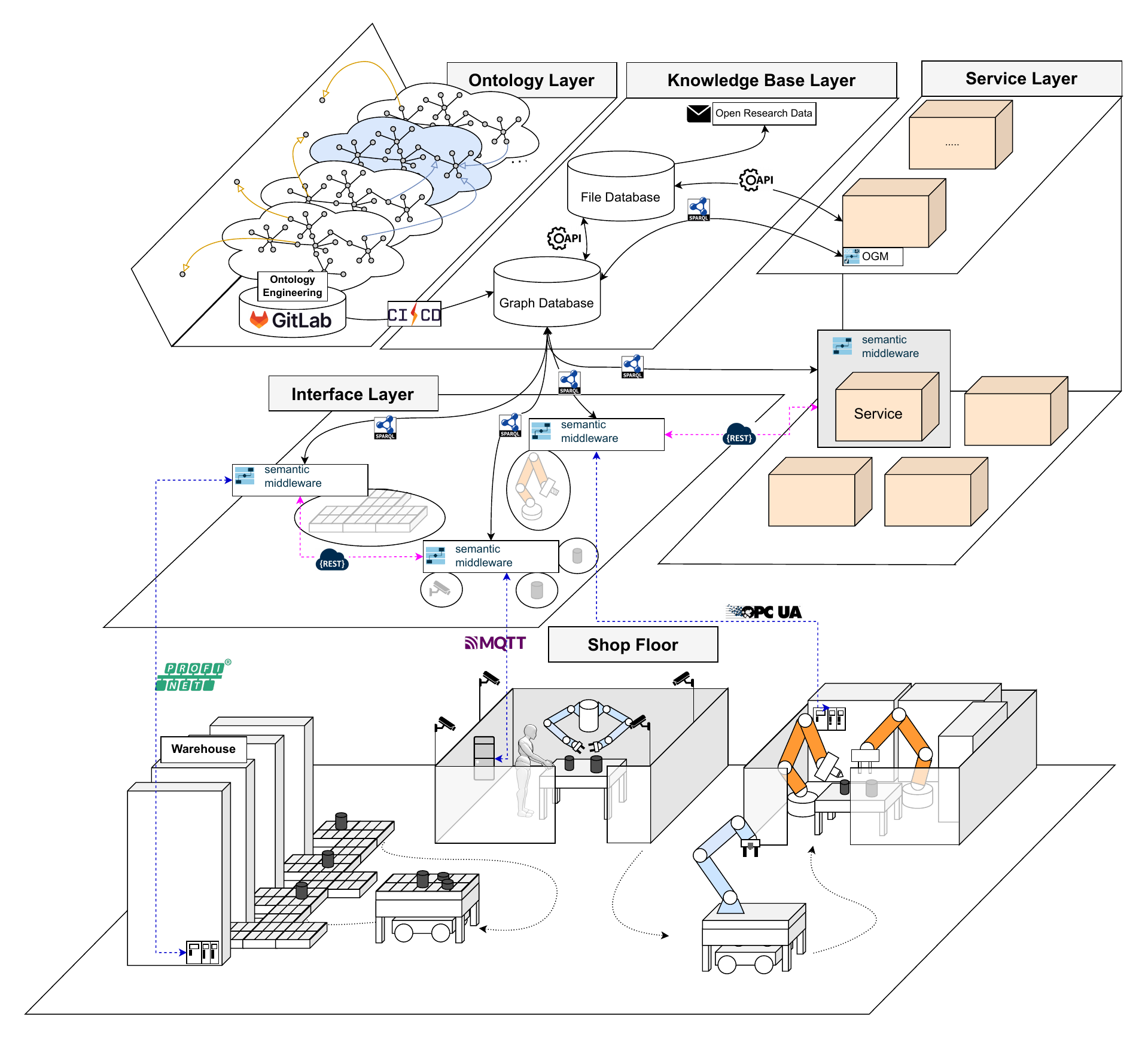}
    \caption{\textbf{KAPPS architecture overview.} The four layers introduced in Sections~\ref{sec:ontologylayer}~--~\ref{sec:interfacelayer} are shown in their integration: the Ontology Layer governs the vocabulary and constraints, persisted into the Knowledge Base Layer through CI/CD pipelines; the Service Layer hosts software services that interact with the graph through OGM-mediated commits; and the Interface Layer connects the upper three layers to physical resources on the Shop Floor through industrial protocols (e.g.: OPC UA, MQTT, ProfiNet). Across all layers, every state-changing operation is mediated by the Knowledge Base, which makes the graph the authoritative runtime representation of the factory information state. SPARQL endpoints and REST interfaces provide the inter-layer transport; protocol-specific connectors absorb the heterogeneity of OT communication at the boundary to the Shop Floor.}
    \label{fig:arch-overview}
\end{figure*}

\noindent 
KAPPS is divided into four layers that collectively realize the requirements derived in Section \ref{sec:perspectives}, as illustrated in Figure \ref{fig:arch-overview}:

\begin{itemize}
    \item The \textbf{Ontology Layer} establishes a shared semantic foundation by defining all system-wide concepts, relations and constraints.
    \item The \textbf{Knowledge Base Layer} provides a unified repository combining semantic knowledge with heterogeneous data persistence, supported by inference and constraint validation.
    \item The \textbf{Service Layer} aggregates and exposes functionalities from distributed systems, including planning, control, detection, annotation, learning, and execution.
    \item The \textbf{Interface Layer} coordinates interactions between heterogeneous shop-floor resources and services, while ensuring consistent synchronization with the Knowledge Base Layer.
\end{itemize}
Collectively, these four layers are realized through a constellation of distributed software components and systems, that jointly enable semantic integration, knowledge-driven decision making and automated event execution within the circular factory. Their architectural conception and concrete implementation are described in the following sections. 

\subsection{Ontology Layer}
\label{sec:ontologylayer}

\begin{figure}[h]
    \centering
    \includegraphics[width=\linewidth]{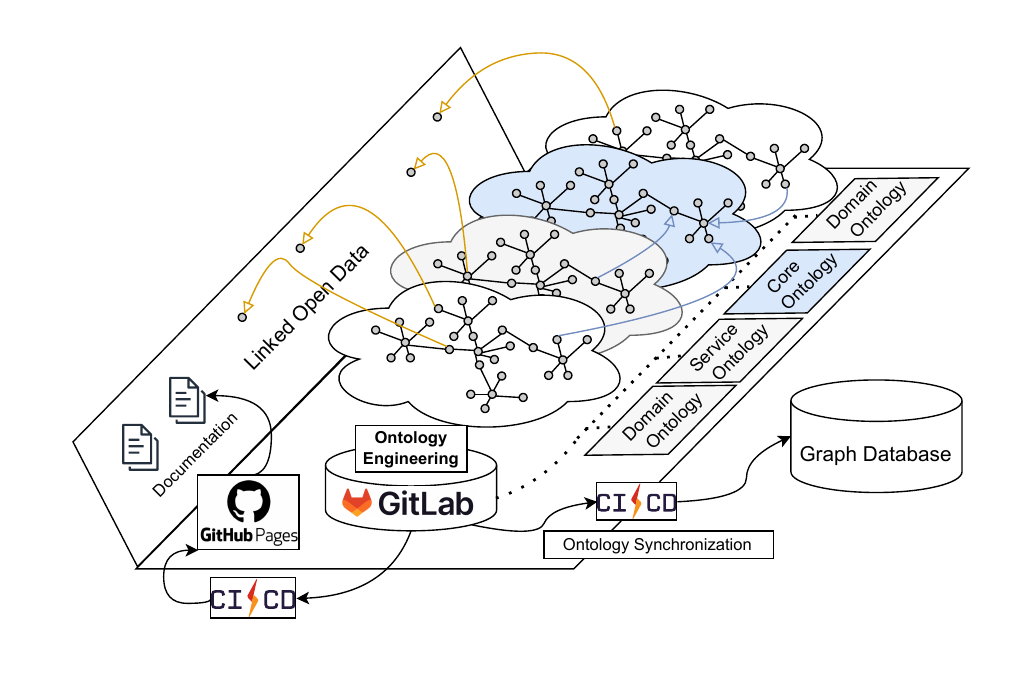}
    \caption{\textbf{Ontology layer: }A core ontology defines system-wide concepts shared across the Circular Factory; a service ontology and multiple domain ontologies extend this core to cover deployment-specific concepts, each maintained as an independent module. External vocabularies from Linked Open Data are referenced where reuse applies. Ontology development is managed under Git-based version control, with a CI/CD pipeline that performs syntactic validation, verifies essential metadata, and synchronizes validated modules into the triple store of the Knowledge Base Layer. Documentation is auto-generated and published for downstream consumers.}
    \label{fig:arch-ontologylayer}
\end{figure}
\noindent 
Ontologies establish the flexible and adaptive data infrastructure required for Circular Factory operation~\cite{hofmannRoleOntologybasedKnowledge2024}. The  ontology layer as illustrated in Figure~\ref{fig:arch-ontologylayer} comprises a core ontology that formalizes the concepts shared across the factory and modular domain ontologies that refine these concepts with domain-specific semantics. The Circular Factories core ontology (cfc:) extends the PPR framework with concepts specific to circular production: a product-variant-generation hierarchy that captures design-time versioning across product generations, a workpiece decomposition into components and assemblies with component origin distinguishing virgin material from recovered cores, temporally evaluated quality and possession states, a capability typology separating equipped, flexibility, and changeability capabilities, and a process typology that formalizes converging, diverging, and unit-preserving production processes~\cite{cfc_core}. The core ontologies documentation is publicly accessible\footnote{\href{https://w3id.org/circularfactory/Core}{https://w3id.org/circularfactory/Core}}, the remainder of this section treats it as a given and focuses on how the KAPPS architecture consumes and operates on it at runtime.

A service ontology extends the ontology layer with a factory-wide standard for exposing distributed functionalities to the knowledge base. It defines two concepts. A service represents a distributed runtime entity wrapped by a single middleware instance. A workflow represents an invokeable function exposed by a service, with preconditions and outcomes that correspond to the arguments and return values of the realizing algorithm. Domain ontologies extend this with concrete service and workflow specializations, contributing one workflow per capability~\cite{pfrommerOntologyRemanufacturingSystems2022}.

Uncertainty in the Circular Factory arises at the level of individual product and process instances and must be represented in a way that preserves instance-specific histories and evolving information states~\cite{heizmannManagingUncertaintyProduct2024}.
The ontology layer accommodates this by treating the representation of an uncertain attribute as a first-class type rather than as a special case: any probabilistic representation that can be expressed as a finite parameter tuple can be declared as an OWL class, with its components, cardinalities, and structural invariants modeled as properties and SHACL constraints in the same way as any other domain concept. 
A concrete and Circular-Factory-specific instance of such a representation is the Gaussian Mixture Model framework proposed by Darijani et al.~\cite{darijaniComprehensiveDescriptionUncertainty2026}, in which each instance-level attribute is held as a parameter tuple that supports closed-form Bayesian fusion of new evidence and admits a directly queryable best estimate via the gaussian fallback property.
The ontology layer hosts such representations alongside scalar and categorical attributes without prescribing the inference operations that consume or produce them; those operations are handled by Service Layer components, as discussed in Section \ref{sec: Servicelayer}.
At the ontology layer, all semantic structures relevant to Circular Factory operation, such as physical constraints, process prerequisites, compatibility relations, lifecycle states, and classification hierarchies are defined declaratively, independent of any storage or execution technology. Their operationalization at runtime is the responsibility of the Knowledge Base and Interface Layers, as discussed in Sections~\ref{sec:knowledgebaselayer} and~\ref{sec:interfacelayer}.

Ontology development is managed using Git-based version control with automated deployment pipelines that perform syntactic validation, verify essential metadata, inject validated modules into the triple store, and auto-generate documentation for downstream developers dynamically to reflect the evolution of the ontology backbone during ongoing ontology engineering.

\subsection{Knowledge base layer}
\label{sec:knowledgebaselayer}
\subsubsection{Unified Knowledge Base}
\noindent 
The Knowledge Base Layer combines a central triple store with specialized databases into a Unified Knowledge Base (UKB) for Circular Factory operation. It serves as the runtime realization of the ontology-defined semantics, hosting a knowledge graph that unifies heterogeneous data, system models, and operational state across the factory environment. The UKB is conceived not as a passive repository but as the operational core of a knowledge-driven CPPS.

The central triple store is currently implemented using GraphDB. it hosts the unified knowledge graph, integrating aligned ontologies together with instantiated data representing concrete products, processes, resources, and their evolving states. By explicitly providing virtual representations of real instances and their relations, the implementation supports instance-level traceability of products, executed operations, resource assignments, and state transitions across the factory lifecycle.

Not all data required in the Circular Factory can be efficiently represented as RDF triples. The architecture therefore follows a multi-database approach: large binary artifacts such as 3D point clouds are stored in databases optimized for unstructured data, while high-frequency sensor streams are persisted in time-series databases. The knowledge graph maintains URI-based references to these external datasets, ensuring semantic contextualization and uniform accessibility while avoiding the performance limitations of direct RDF storage.

For managing research and experimental data, the Kadi ecosystem~\cite{brandtKadi4MatResearchData2021} is employed as a specialized file and metadata repository. Kadi provides persistent storage for heterogeneous artifacts, flexible metadata management, FAIR-compliant dissemination, and a graphical interface for human interaction. Semantic interpretation and constraint enforcement remain within the triple store, while Kadi records are semantically interlinked with corresponding graph entities, enabling a unified view over currently active PPR Instances on the shop floor and the interlinked PPR-Types, including Type specific process and design knowledge, specifications, and research data.

\subsubsection{Constraint Enforcement }
\noindent 
To operationalize the semantics defined at the ontology layer, the Knowledge Base Layer applies both OWL reasoning and SHACL-based validation to the instantiated knowledge graph. These mechanisms address complementary aspects of semantic consistency.

OWL reasoning derives implicit knowledge from explicitly asserted facts -- such as inferred classifications, capability dependencies, and resource–process compatibilities -- under an open-world assumption. This supports semantic integration and information fusion across heterogeneous data sources without requiring all relevant facts to be explicitly stated.

SHACL validation~\cite{shacl2017} provides closed-world constraint checking over concrete system states. Unlike OWL reasoning, which enriches the graph with inferred knowledge, SHACL enforces that specific conditions hold and rejects updates that would violate them. In the KAPPS architecture, SHACL shapes encode relational and cross-entity constraints that govern physical state transitions. For example, that a Product cannot be simultaneously possessed by two resources in different locations, or that an operation cannot commence until prerequisite conditions are met. These constraints cannot be evaluated at the level of individual entity projections; they require visibility into the complete graph state.

The KAPPS architecture enforces constraints at two complementary layers, reflecting the inherent incompleteness of any partial projection of the knowledge graph into a runtime object. At the Interface Layer, the Object–Graph Mapper derives validators from OWL class definitions, enforcing single-entity constraints such as property types or maximum cardinalities at the moment of object materialization. These validators operate on isolated entity projections and cannot evaluate constraints referencing other entities. Relational constraints governing interactions between multiple instances are enforced at the Knowledge Base Layer by the triple store's native SHACL engine. Because the triple store maintains the complete graph, SHACL shapes can express closed-world constraints over arbitrary triple patterns, rejecting any update that would violate cross-entity consistency before it is admitted into the persistent store.
The current implementation uses GraphDB as the triple store backend, but the used SHACL shapes do not rely on any vendor-specific extensions. This ensures portability: any triple store providing a conformant SHACL validation engine can serve as the constraint enforcement backend without modification to the shapes or the integration layer.

The separation between single-entity and cross-entity validation ensures that structural violations detectable from a single entity projection are caught decentrally and early in the write pipeline, while relational violations requiring cross-entity reasoning are caught at persistence time by the triple store's SHACL engine. Both classes surface to the caller at the same API boundary (the commit call) as typed exceptions carrying the validation report, differing only in the constraint class they report on and the layer that raised them.
Application-level parsing of validation reports for differentiated error handling or automated corrective action is designated for future work. Section~\ref{sec:evaluation} demonstrates this enforcement mechanism through a concrete use case.
\vspace{0.5cm}

Decision provenance and temporal reproducibility are handled by two distinct mechanisms at this layer. Provenance is carried by the ontology itself: every observation, measurement, and state change can reference a PROV~\cite{w3c_prov_o_2013} qualified description of its originating activity, and the OGM enforces this requirement as part of the validators it derives from the ontology, so that no provenance-unqualified fact can enter the graph.

Temporal reproducibility is carried by the triple store: any history tracker that records changes at triple granularity can be deployed to handle versioning, and from such a record any past system state is directly reconstructible. The demonstrator deploys the GraphDB history plugin in this capacity; the architecture commits to no specific implementation.

\subsection{Service layer}
\label{sec: Servicelayer}
\noindent 
The Service Layer hosts the distributed services that collectively constitute the factory's control logic. Its design follows the principles of service-oriented architecture for manufacturing, whose suitability for production planning and control has been established through systematic review against the criteria of discoverability, modularity, interoperability, loose coupling, location transparency, and composability~\cite{Behrendt2023}.

KAPPS organizes this layer through two concepts. A \emph{service} is a distributed runtime entity wrapped by a single middleware instance: a planner, controller, scheduler, learner, or annotation component running as an addressable unit on the network. A \emph{workflow} is an invokeable function exposed by a service, with preconditions and outcomes corresponding to the arguments and return values of the realizing algorithm. Each service exposes one or more workflows, and each workflow realizes one operational capability of the PPR framework~\cite{pfrommerOntologyRemanufacturingSystems2022}: the callable function through which a capability such as \emph{move} or \emph{unscrew} is triggered on a concrete resource.

This convention follows the microservices sense of service rather than the capability–skill–service (CSS) notion established in the AAS tradition~\cite{kocherReferenceModelCommon2023a}, where ``service'' denotes the invocation interface of a single capability or skill. In KAPPS that role is occupied by a workflow; the term service is reserved for the deployable, middleware-wrapped entity hosting one or more such workflows.

Services and their exposed workflows interact with the Unified Knowledge Base through two asymmetric paths, both mediated by Interface-Layer components. Reads may be issued either through the OGM or directly through the triple store access module, while every write originates as an OGM commit. The access module exposes no SPARQL UPDATE interface, making the OGM the architecture's single validated write path. Services consume validated semantic state, derive decisions from it, and commit outcomes back through this single write path, remaining independent of concrete database technologies and communication protocols.

The semantic middleware manages the lifecycle of each service as a wrapper around its domain-specific logic. On startup, the middleware instantiates the service individual together with the workflow individuals it exposes in the knowledge graph according to the service ontology introduced in Section~\ref{sec:ontologylayer}, links the service to the providing resource, and populates its address property with the endpoint at which its workflows are reachable. While the service is running, its workflows are discoverable through SPARQL queries over the graph; on shutdown, the middleware removes the address property so that no workflow exposed by the service appears as invokeable, while preserving the service and workflow individuals themselves. This separation of availability from existence serves two purposes: it reflects the reality of a changeable production system in which a capability may be physically present but temporarily offline, and it supports provenance: the knowledge graph retains a record of which services have existed and which workflows they have exposed, independently of their current operational status.

This mechanism gives planning services a live view of the operational capabilities currently available in the factory. An operation can be modeled as an object that materializes a task-capability match by referencing a concrete workflow individual: the planning layer queries for services providing the required capability, retrieves the workflow endpoint from the graph, and invokes the operation. Planning logic is decoupled from deployment topology and adapts to services coming online or going offline without reconfiguration. Section~\ref{sec:evaluation} demonstrates this mechanism in the FlexConveyor use case, where conveyor modules register their capabilities and respond to semantically mediated operation requests.

Human decisions are treated as conceptually equivalent to automated workflows: operator inputs, expert assessments, and manual interventions are abstracted as outputs that update the knowledge graph in the same manner as algorithmic decisions. This uniform treatment enables consistent provenance tracking and constraint validation across all decision sources. The internal algorithms of individual workflows are domain-specific and outside the scope of this work; the focus here is on the integration infrastructure that enables coherent operation on a shared semantic representation.

\begin{figure}
    \centering
    \includegraphics[width= \linewidth]{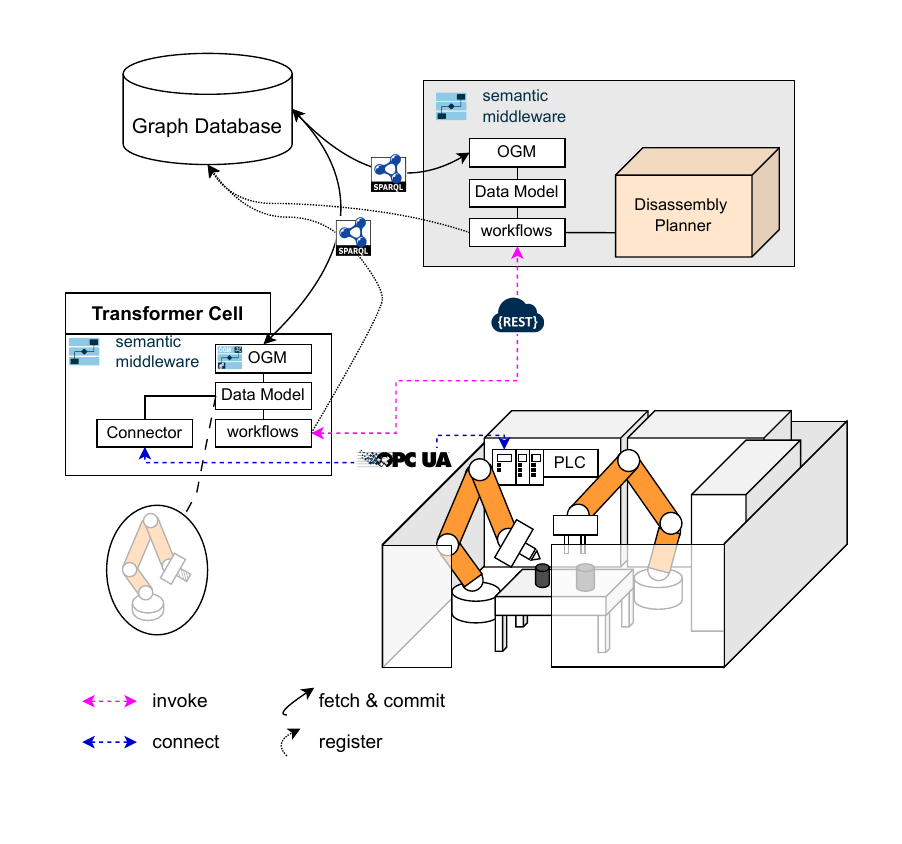}
    \caption{ \textbf{Interface and service layer in operation.}Two middleware instances coordinate through the graph database: the left instance wraps a Transformer Cell and bridges industrial hardware via an OPC UA connector; the right instance wraps a pure software service (here, a disassembly planner). Both register their workflows as typed service individuals in the graph at startup, exchange data through the OGM (fetch \& commit), and invoke peer workflows directly via REST once discovered through SPARQL. The connect action handles the OT crossing where the middleware meets protocol-specific hardware interfaces.}
    \label{fig:arch_serviceinterface}
\end{figure}

\subsection{Interface Layer}
\label{sec:interfacelayer}
\noindent 
The Interface Layer constitutes the integration stack of the Circular Factory, providing the runtime bridge between ontology-grounded knowledge representations, distributed services, and physical shop floor resources. Rather than forming a monolithic middleware, it is architected as a modular composition of three components that together enable consistent interaction with the Unified Knowledge Base across heterogeneous IT and OT environments. The \emph{triple store access module} provides the foundational communication layer with the triple store, abstracting triple store-specific protocols behind a uniform API. Building on this foundation, the \emph{Object–Graph Mapper (OGM)} establishes the  bridge between the knowledge graph and object-oriented runtime environments, enabling services to consume and modify instance representations from the knowledge base through validated data objects. Finally, the optional \emph{semantic middleware runtime} extends these capabilities with event-driven coordination, protocol integration, and distributed deployment modes for shop floor resources. These components can be used independently or in combination, depending on integration requirements: lightweight,  asynchronous services may interact directly through the OGM, while complex coordination scenarios leverage the full middleware stack.

\subsubsection{Triple store Access Module}
\noindent 
At the foundation of the Interface Layer, the triple store access module provides programmatic communication with the semantic backend. It exposes graph operations such as triple retrieval, insertion, update, deletion, and existence checks through a uniform API, together with utilities including IRI validation, structured triple representations, and XSD-to-Python datatype mappings. The current implementation\footnote{\href{https://github.com/JaFeKl/graph\_db\_interface}{https://github.com/JaFeKl/graph\_db\_interface}}  targets GraphDB via its REST API, but backend-specific access is confined to a single adapter; all higher-level functionality operates on standard SPARQL and RDF abstractions, permitting alternative triple stores to be integrated with minimal modification.

\subsubsection{Object–Graph Mapper (OGM)}
\noindent 
The Object–Graph Mapper (OGM)\footnote{\href{https://github.com/SAWeindel/kapps\_ogm}{https://github.com/SAWeindel/kapps\_ogm}} establishes the bridge between the ontology-based knowledge graph and object-oriented runtime environments. Unlike compile-time code generators such as on2ts~\cite{wrightOn2tsTypescriptGeneration2020} or MetaFactory~\cite{xiaoMetaFactoryCloudbasedFramework2025} reviewed in Section~\ref{sec:review}, the KAPPS OGM derives type-safe class schema dynamically from the live triple store at runtime. There is no code-generation step, and an ontology update propagates to every service on its next fetch without redeployment. The object-derivation principle itself is not unprecedented: Owlready2~\cite{lamyOwlreadyOntologyorientedProgramming2017}, for example, demonstrates that OWL ontologies can be loaded directly as Python classes and accessed through native attribute syntax. Owlready2 is, however, designed as a single-process ontology engineering library that maintains its own local quad store, not as a runtime layer sitting between distributed services and a shared, authoritative external triple store. The KAPPS OGM extends runtime ontology-derivation into that distributed CPPS setting and combines it with type-safe materialization through Pydantic validators automatically derived from OWL semantics: property domain and range determine which properties may be attached to an object based on its class hierarchy (a property is admissible on any subclass instance of its declared domain), and OWL property restrictions encode further per-class shape constraints such as maximum cardinalities, datatype constraints, and required-attribute declarations so that ontology violations are caught at the Python boundary before any SPARQL update is constructed. The ontology is in this sense not a documentation artifact or design-time schema description: it is the live, authoritative declaration of every class, property, and constraint that runtime services see, and remains so for as long as the system is running.
When a client requests an entity of a given class, the mapper issues SPARQL queries to retrieve the class definition, property specifications, cardinality constraints, and subclass hierarchy directly from the currently deployed ontology, transforming them into Pydantic model classes with corresponding field types and validators. The mapper supports create, read, and update operations through this unified object interface: instances fetched from the knowledge graph are materialized as validated Python objects, and modifications are persisted back to the triple store via automatically generated SPARQL patterns. Entity-level deletion is intentionally unsupported (products, processes, and resources recorded in the knowledge graph must be retained for traceability and learning) but at the triple level, full mutability is possible: individual property assertions can be added, modified, or removed as operational state evolves. For example, when a service becomes unavailable, automated handlers remove its accessibility specifications while preserving the service entity and its historical interactions.
To ensure efficient interaction with large ontologies, instance materialization follows a demand-driven expansion strategy. Only explicitly requested subgraphs are resolved into concrete objects; all other entities are retained as IRI references, that can be dynamically expanded later on. Services control this expansion through scope specifications, obtaining precisely the projection required for their task while remaining consistent with the Unified Knowledge Base. The Pydantic validators derived at materialization time provide early detection of malformed data with detailed Python-level diagnostics, complementing the cross-entity SHACL validation performed at the Knowledge Base Layer.
Because the OGM derives runtime types directly from the live ontology, attributes of arbitrary structural complexity require no special handling. A scalar float and a Gaussian Mixture Model parameter tuple in the sense of Darijani et al.~\cite{darijaniComprehensiveDescriptionUncertainty2026} pass through the same dynamic schema-derivation path, are materialized as the same kind of validated Pydantic object, are persisted through the same SPARQL UPDATE generation, and are exposed through the same fetch/commit interface. Service-Layer components implementing the corresponding inference operations -- Bayesian fusion of new evidence, conditioning on observed measurements, marginalization via the closed-form expectation of the mixture -- consume and produce these tuples through the standard OGM interface. The architectural substrate for instance-level uncertainty representation is therefore provided structurally by the OGM, while the inference content is contributed by complementary work and integrated as ordinary Service-Layer logic.

\subsubsection{Semantic Middleware Runtime}
\noindent 
Operational coordination across services is realized through the semantic middleware runtime, which provides service orchestration, a connector framework for protocol mediation, and a REST engine for workflow invocation. The implementation is a refactor of the aas-middleware framework~\cite{behrendtAasmiddlewareEnablingInteroperable2025}: the Asset Administration Shell data model, the persistence layer, and the AAS-specific logic have been removed entirely, retaining only the orchestration, connector, and REST components. The in-memory data model held by each middleware instance is provided by the OGM, so services operate on data objects that reflect the ontology-defined schema while benefiting from the middleware's coordination capabilities. The runtime can be deployed either as a centralized service or in a decentralized peer-to-peer configuration on individual resources.

The middleware routes messages between heterogeneous endpoints. The OGM determines message structure derived from the ontology, while the middleware determines message destination: other middleware instances via REST, industrial endpoints, external databases, or the knowledge graph. Domain services therefore operate on data objects that directly conform to the ontology-defined structure, without managing communication protocols directly. Protocol and database integration is realized through a uniform connector interface requiring the four methods \emph{connect}, \emph{disconnect}, \emph{provide}, and \emph{consume} without prescribing their internal implementation, so that any client implementation handling messaging over an arbitrary protocol can act as a connector and industrial protocols, databases, and inter-service communication are handled through a single abstraction. Synchronization with the knowledge graph is itself a connector in this framework: an upstream connector wraps the OGM's operations and persists changes through explicit commit calls triggered by service logic at semantically meaningful state transitions, mirroring the transaction patterns of conventional manufacturing execution systems.

\subsubsection{Conclusion}
\label{sec:architecture-conclusion}
\noindent 
The four layers cooperate to realize instance-level Product--Process--Resource traceability within a unified semantic infrastructure. Products entering the Circular Factory are instantiated as knowledge graph entities with persistent identifiers; as they traverse the shop floor, each operation, observation, resource assignment, and state transition is recorded as semantic relations linking the product instance to process instances, resource instances, and temporal contexts. The multi-database architecture keeps high-volume sensor data and binary artifacts semantically linked without overloading the triple store. The OGM routes every service write through a single validated path and provides runtime access to the resulting instance graphs, while SHACL shapes enforce that recorded state transitions satisfy physical and operational constraints. The result is a queryable, validated, and complete record of each product's lifecycle that supports both real-time control and retrospective learning.

This is the composition that Section~\ref{sec:review} identified as missing from the state of the art. The reviewed knowledge-graph approaches contribute semantic expressiveness but stop at planning-time analysis; the reviewed AAS-based approaches contribute service integration, protocol mediation, and event-driven execution, but their hierarchical, per-asset submodel structure provides no native mechanism for inter-entity relational validation and treats the data layer as a passive store rather than as a constraint enforcement layer. The KAPPS architecture combines both classes of capability around a different center of gravity: an ontology-grounded knowledge graph operated as the authoritative write-time layer through which every execution-time operation passes, with operational coordination, protocol mediation, and event-driven execution realized against the graph rather than against the AAS submodel. The following section demonstrates this architecture through two implemented use cases.

\FloatBarrier
\section{Demonstration }
\label{sec:demonstration}

\subsection{Use Case 1: Anomaly Detection and Learning Through Knowledge-Graph-Mediated Service Integration}
\label{sec:uc1}

\begin{figure}[tb]
    \centering
    \includegraphics[width=\linewidth]{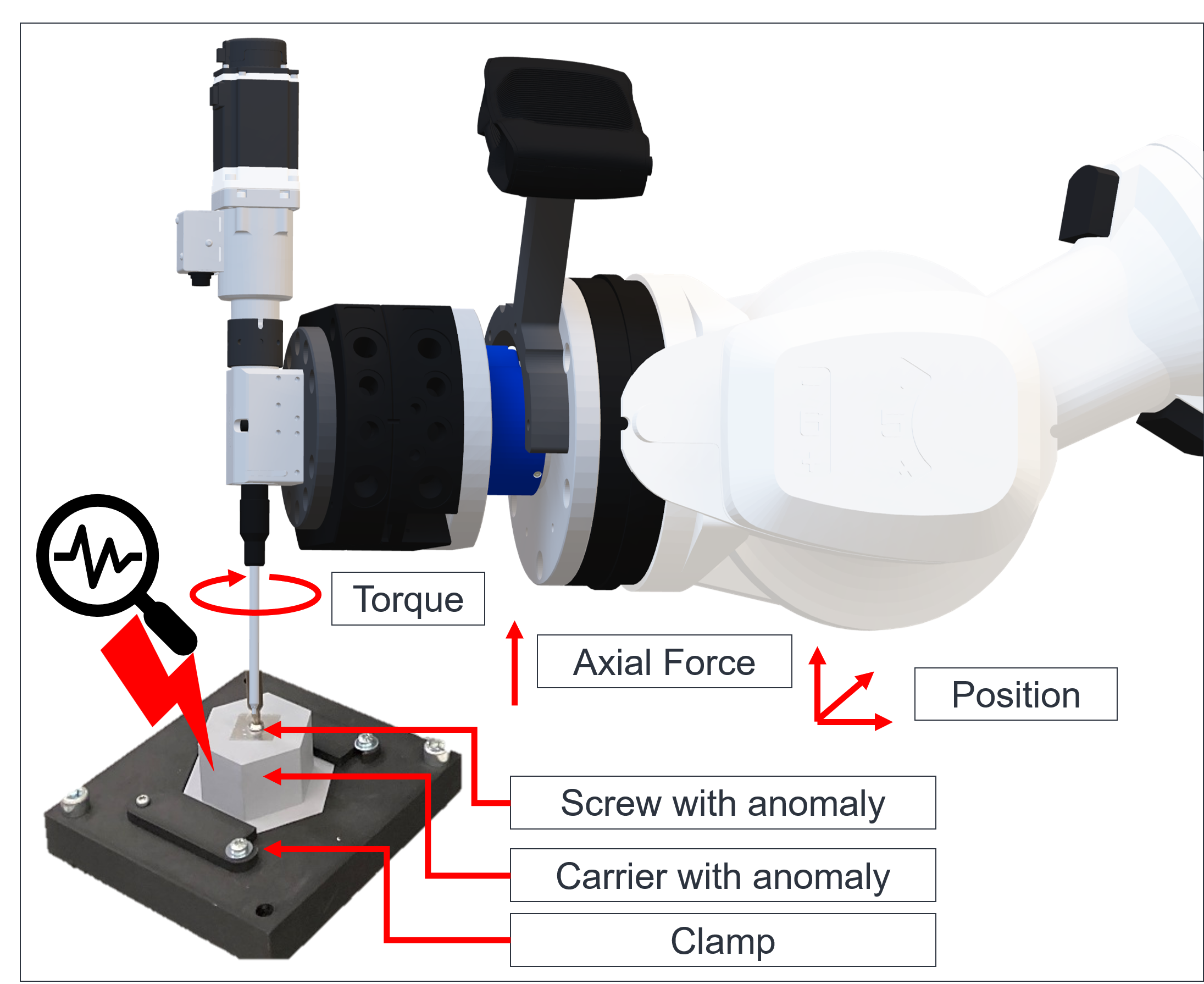}
    \caption{Physical setup of the robotic transformer cell.
    The screwing resource mounts a screwdriver on a six-axis robot arm.
    Torque is measured at the screwdriver; axial force and TCP position
    are provided by a force-torque sensor and the robot controller,
    respectively.
    In the real cell, ProfiNet carries the motion state and OPC~UA
    exposes the force-torque readings; in the demonstrator, both
    channels are mocked by a knowledge-graph-backed replay service.}
    \label{fig:tfcell_experiment_setup}
\end{figure}
\noindent 
The gap analysis in Section~\ref{sec:review} established that existing knowledge-based CPPS approaches treat the knowledge graph as a read-oriented repository: a planning-time artifact that is queried before execution begins, rather than an active integration substrate mediating execution as it unfolds.
This use case demonstrates how KAPPS addresses that gap in a concrete Circular Factory scenario: the robotic unscrewing of screws from recovered angle grinders undergoing disassembly.
The angle grinders product type model is described in detail in~\cite{VlajicMartinFischer2025}, and their removal is performed by a robotic transformer cell, equipped with a screwing resource. Due to substantial variance in screw condition, the unscrewing process fails regularly; failures must therefore be detected in-process to minimize downtime.

The demonstrator integrates three services that operate at different abstraction levels and communication interfaces, a perception service, an anomaly detection (AD) service, and a learning service, whose coordination is mediated exclusively through the knowledge graph, realizing an information-centric architecture in which heterogeneous sensor streams, classification decisions, and learned parameter updates are all represented as first-class semantic entities.
The anomaly detection methodology itself will be detailed in a future publication; this section focuses on the architectural
integration that enables it.

\subsubsection*{Physical Setup and Sensing}
\noindent 
The transformer cell employs a screwing resource that consists of an electric screwdriver, mounted on a six-axis robot arm \ref{fig:tfcell_experiment_setup}. During each unscrewing attempt, the perception service acquires three concurrent sensor streams: the torque $M_y$ applied to the screw axis, the axial force $F_y$ along this screw axis, and the TCP position $P_y$ along it.
In the physical cell, these signals are delivered via ProfiNet carrying the robot controller's motion state and force-torque sensor readings. The signals are ingested by the perception service through corresponding middleware connectors \ref{fig:uc1_arch}.
To permit reproducibility without physical hardware, both protocol connectors are replaced in the deployed\footnote{\href{https://github.com/EHoffm/CPPS\_Circular\_factory\_usecases\_JMS/}{https://github.com/EHoffm/CPPS\_Circular\_Factory\_usecases\_JMS/}} demonstrator by a knowledge-graph-backed mock that serves pre-recorded time-series from files. All downstream services are unaffected by this substitution because they interact exclusively with the knowledge graph, not with the protocol layer directly.

\subsubsection*{The Closed-Loop Architecture}

\begin{figure}[tb]
  \centering
  \includegraphics[width=\linewidth]{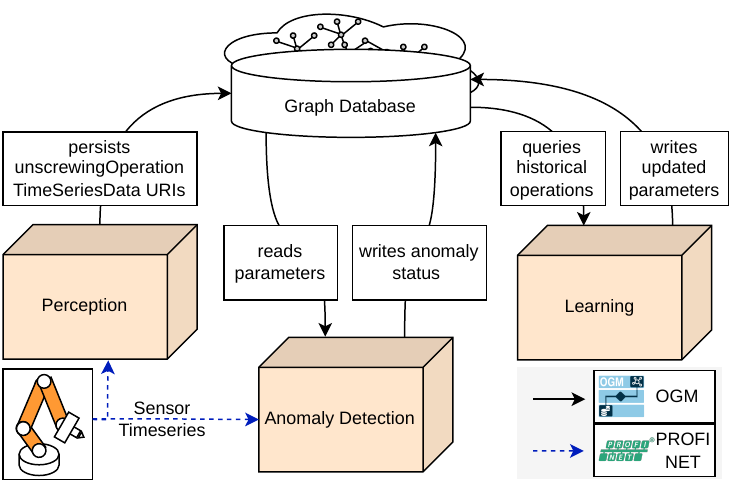}
  \caption{Closed-loop architecture of UC1.
    The knowledge graph is the sole shared medium between the three
    services.
    The perception service persists multimodal sensor references
    (\texttt{TimeSeriesData} URIs for torque, axial force, and
    position) as \texttt{UnscrewingOperation} individuals.
    The anomaly detection service reads type-level detection parameters
    from the knowledge graph and writes anomaly decisions co-localized
    with the active parameter snapshot.
    The learning service derives updated parameters from accumulated
    operation histories and commits them back.
    InfluxDB stores the raw time-series; the knowledge graph holds
    semantically navigable URI references.}
  \label{fig:uc1_arch}
\end{figure}
\noindent 
Figure~\ref{fig:uc1_arch} shows the three services and the data flows
between them, with the knowledge graph as the sole shared medium.
After each unscrewing attempt, the perception service persists a new
\texttt{:UnscrewingOperation} instance in the knowledge graph.
This instance carries three \texttt{:TimeSeriesData} object-property
links, one each for the torque, axial-force, and position
streams, whose \texttt{:hasJSONEncodedTimeSeriesData} properties encode
URI references to the corresponding high-frequency records in
InfluxDB.
Persisting URI references rather than raw samples keeps the graph database
unencumbered by high-frequency data while maintaining
semantically navigable links from every operation instance to its
underlying measurements (R2, R5).

The AD service reads the current detection parameters for the relevant
screw type directly from the knowledge graph via the
OGM's \texttt{fetch} operation at the start of each cycle.
These parameters, are typed datatype properties on the
screw type instance and encode the thresholds used by a
sequential decision logic that iterates over the three sensor streams
simultaneously to classify the attempt.
Detectable failure modes include missing screws, occluded or rounded
screw heads, loose anchors, and stuck screws.
Upon classification, the AD service writes the resulting anomaly label back to the operation instance as datatype properties.
This co-localisation of the decision outcome and the decision context in a
single semantic record constitutes the decision-level provenance
required by R12: it is always possible to recover, for any past
operation, parameters where used when the
classification was made, and to trace those parameters back to the
learning episode that produced them.

The learning service closes the loop.
It periodically queries the knowledge graph for all completed
\texttt{:UnscrewingOperation} instances associated with a given screw
type, follows the \texttt{:TimeSeriesData} links to retrieve the
corresponding measurement batches from InfluxDB, and applies
regression-based algorithms to derive updated estimates of the five
detection parameters.
The updated values are written back to the screw type instance via the
OGM's \texttt{commit} operation. The AD service obtains them on its next
\texttt{fetch} cycle.
Each iteration of this loop extends the system's operational
experience: the learned parameter values are not held in an external
model store that is disconnected from the rest of the factory's information
but are persisted as typed ontology properties on the same screw type
instance that planning and execution services query at runtime
(R13).

\subsubsection*{Ontological Modeling}
\begin{figure*}[tb]
  \centering
  \includegraphics[width=1\linewidth]{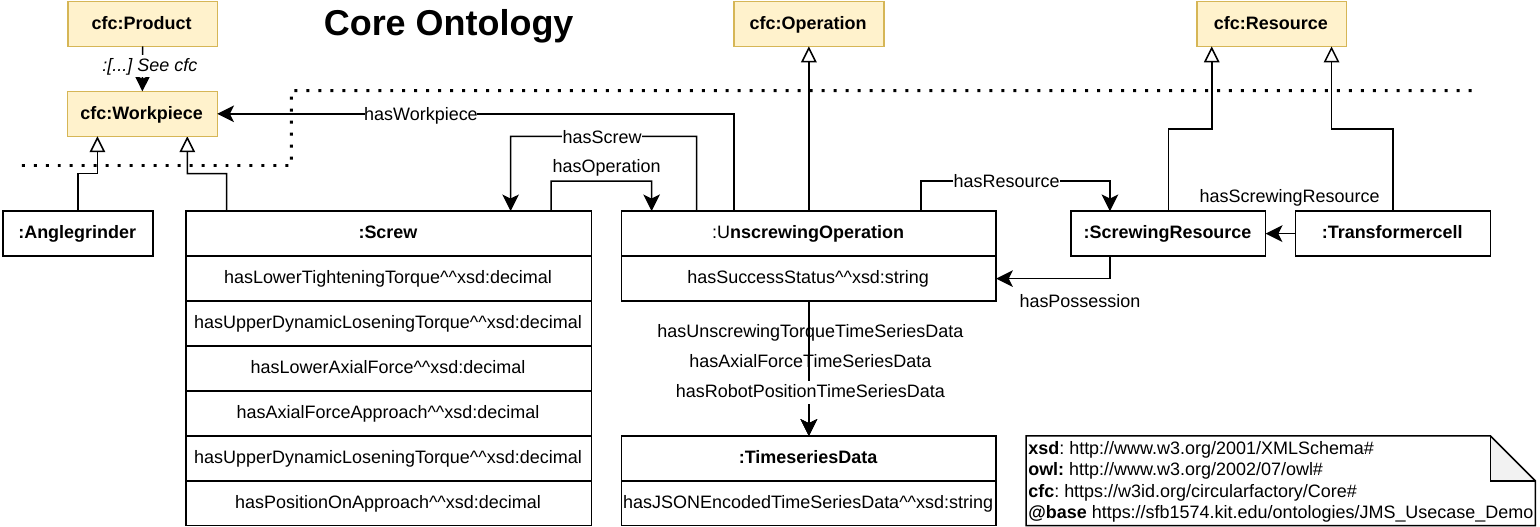}
  \caption{Minimal domain ontology for the anomaly detection -learning loop. Because the KAPPS infrastructure absorbs OGM binding and middleware connectivity, the application-level model reduces to three main classes and their interrelations: a \texttt{:Screw} carrying detection parameters, an \texttt{:UnscrewingOperation} linking sensor time series to anomaly decisions, and a \texttt{:ScrewingResource} for PPR traceability.}
  \label{fig:uc1_ontology}
\end{figure*}
\noindent 
Figure~\ref{fig:uc1_ontology} shows the ontology for this use
case.
The \texttt{Screw} class is defined as a subclass of \texttt{:Product}
and carries the five detection parameters as typed datatype properties
at the type level.
A more detailed description of the product model into which
the screw is embedded can be found in~\cite{VlajicMartinFischer2025, cfc_perpetual_product} \footnote{\href{https://w3id.org/circularfactory/PerpetualProduct}{https://w3id.org/circularfactory/PerpetualProduct}}(R6).

The \texttt{:UnscrewingOperation} has an anomaly decision accessible via \texttt{:hasSuccessStatus},
relates to the \texttt{:Screw} it was executed on, the executing
\texttt{:ScrewingResource}, and the three \texttt{:TimeSeriesData}
sensor records.
\texttt{:Screw}, \texttt{:UnscrewingOperation}, and
\texttt{:ScrewingResource} are linked bidirectionally to provide
navigable PPR traceability at the instance level~(R4).

None of the three services writes SPARQL or interprets RDF directly.
All three interact with the knowledge graph through the OGM-validated
object interface: the perception service uses \texttt{create} and
\texttt{commit} to instantiate new operation individuals; the AD
service uses \texttt{fetch} to obtain typed Pydantic \footnote{\href{https://pydantic.dev/docs/validation/latest/get-started}{https://pydantic.dev/docs/validation/}} objects whose
fields correspond to the ontology-defined parameters; and the learning
service uses \texttt{fetch} to retrieve screw type instances with their
associated operation and measurement links, and \texttt{commit} to
persist the updated parameters.
Any write that would violate the ontology-derived schema is rejected
before reaching the graph database, ensuring that the information state
remains structurally valid throughout execution (R1).

\subsubsection*{Traceability Chain}
\noindent 
The instance-level traceability produced by this architecture can be
followed concretely.
A screw instance \texttt{ex:Screw\_4711}, belonging to the angle-grinder
product \texttt{ex:AngleGrinder\_099}, is linked via \texttt{:hasScrew}
to the operation instance \texttt{ex:UnscrewOp\_2024-03-15T14:32}
executed by \texttt{ex:Screwing\-Resource\_1} within
\texttt{ex:TransformerCell\_A}.
That operation instance carries three \texttt{Time\-Series\-Data} links
whose URI properties reference the torque, axial-force, and
position records in InfluxDB, the same multimodal streams (R2)
that the AD service classified against the parameters
$M_{y,\mathrm{lower}} = \SI{0.1}{\newton\metre}$ and
$M_{y,\mathrm{upper}} = \SI{10}{\newton\metre}$ that were active on
\texttt{ex:Screw\_4711}'s type at that moment.
The \texttt{:hasSuccessStatus} property on the operation instance records the resulting classification, co-localized with the parameter snapshot that produced it (R5). Because the \texttt{:UnscrewingOperation} instance links to both the three \texttt{:TimeSeriesData} records and, via \texttt{:hasScrew}, to the \texttt{:Screw} type instance carrying the detection parameters active at the time of classification, every past decision is fully reconstructible from the knowledge graph alone (R12).
The detection parameters active at classification time are themselves learned abstractions persisted as typed datatype properties on the \texttt{:Screw} type instance. This makes them query-able by any service without knowledge of the learning procedure that produced them (R13).

\subsubsection*{Contribution}
\noindent 
This use case demonstrates, in complementary fashion to
Section~\ref{sec:uc2}, a distinct set of architectural capabilities.
Where UC2 demonstrates SHACL-based constraint enforcement and
decentralized coordination across peer services at runtime, UC1
demonstrates the OGM as a bidirectional integration mechanism for
heterogeneous subsystems operating on physical hardware, the knowledge
graph as the active backbone of a closed perception--detection--learning
loop, multi-database integration with semantically navigable links to
external time-series storage, and instance-level PPR traceability
linking decisions to their underlying multimodal sensor evidence.

\begin{figure*}[ht]
    \centering
    \includegraphics[width=1\linewidth]{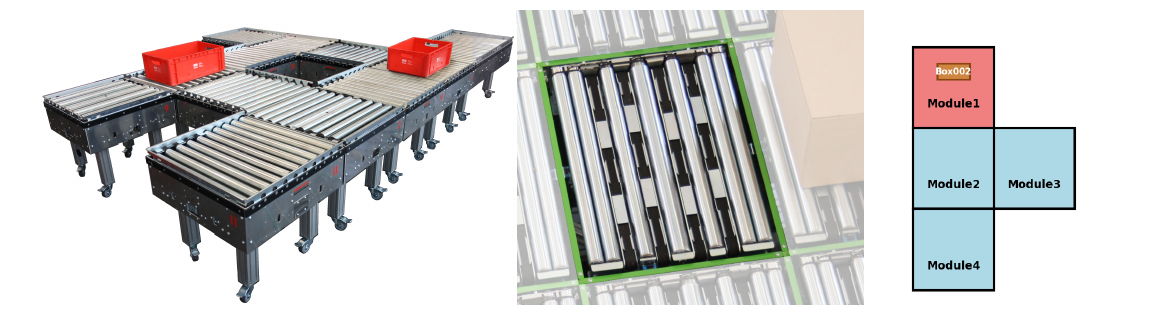}
    \caption{The FlexConveyor System~\cite{Mayer2009_1000011463} serving as use case example. From left to right: system overview, the conveying unit allowing conveyance into four cardinal directions, and the abstract system visualization provided in the graphical interface to monitor system state.}
    \label{fig:Flexconveyor}
\end{figure*}
\subsection{Use Case 2: A Decentralized Conveyor Network Coordinated through a Constraint-Enforced Knowledge Graph} \label{sec:uc2}
\noindent 
A property of a CPPS following the KAPPS Architecture is that resource and service coordination can be realized over a constraint-enforced knowledge graph rather than over a centralized controller. This use case demonstrates that property concretely: a network of autonomous conveyor modules coordinate box transfers through a shared knowledge graph, while the graph itself enforces the physical invariants that make distributed coordination safe.

As a demonstrator serves the distributed FlexConveyor system as introduced by~\cite{Mayer2009_1000011463}, shown in Figure~\ref{fig:Flexconveyor}: modular conveyor units, each capable of holding at most one transport box at a time and of conveying boxes in the four cardinal directions. Modules operate autonomously, initiating convey operations based on local routing decisions in the sense of~\cite{Mayer2009_1000011463} without centralized state visibility, so that no single module controller holds a consistent view of which module possesses which box at any moment. A simplified warehouse management system (WMS) connects the modules to a production context by creating boxes with assigned origin and destination modules, serving as an abstraction of the external production environment.

In a setting of this kind, preventing invalid system states is a precondition for reliable operation. The review in Section~\ref{sec:review} established that existing applications of SHACL in manufacturing validate static structural properties~\cite{wagnerValidatingSystemBehavior2025}, pre-task capability parameters~\cite{davidDeployingOWLOntologies2023}, or scalar sensor value ranges~\cite{morenoSemanticAssetAdministration2024}, but none enforces the feasibility of state transitions across interrelated product and resource instances during production operation. The remainder of this section shows how distributed coordination is realized through event-driven workflows over the graph, and how SHACL enforcement at the persistence layer prevents control flows that would create contradictions between the physical system and its digital representation.

The FlexConveyor domain is captured in the ontology layer through three commitments. Modules are typed as \texttt{:FlexConveyorModule}, a subclass of \texttt{cfc:Resource}; boxes are instances of \texttt{:Box} that progress through three lifecycle states: \texttt{Created}, \texttt{InTransit}, and \texttt{Delivered} are linked via \texttt{:hasState}; and the handover between modules is carried by a possession relation \texttt{:hasPossession} (domain \texttt{cfc:Resource}, range \texttt{:Box}) with its inverse \texttt{:isPossessedBy}, so that each box is traceable to the module currently responsible for it. Each module's coordination interface is modeled by three workflows: \texttt{:ConveyWorkflow}, \texttt{:ReserveWorkflow}, and \texttt{:ReceiveWorkflow}, each instantiated when a module comes online and registered to the graph by the middleware. The physical invariants that this type model must respect are formalized as SHACL shapes deployed in the triple stores shapes graph. Structural constraints use SHACL core vocabulary; state-dependent cross-entity constraints use SHACL-SPARQL to evaluate conditions spanning both the box's lifecycle state and the resource's possession relation. Listing~\ref{lst:shacl-intransit} shows \texttt{:InTransitBoxPossessedShape} as a representative example; the complete shape set is available in the associated repository\footnote{\href{https://github.com/EHoffm/CPPS\_Circular\_factory\_usecases\_JMS/}{https://github.com/EHoffm/CPPS\_Circular\_factory\_usecases\_JMS/}}.

Communication between modules and with the WMS is handled via the REST API of each middleware instance. The three workflows together implement a three-step handshake for box transfer: the sending module requests a reservation on the next hop through \texttt{:ReserveWorkflow}; once confirmed, the receiving module triggers conveyance on the source through \texttt{:ConveyWorkflow}. During conveyance, possession of the box transfers from the sending to the receiving module following the handover principle of~\cite{ElfahamEpple+2020+208+221}, and arrival is acknowledged through \texttt{:ReceiveWorkflow}. The receive step abstracts the physical recognition of the arriving box, allowing the demonstrator to run as simulation without FlexConveyor hardware. A graphical interface bootstraps the module topology and monitors box locations as a read-only observer of the knowledge graph; it does not modify system state during operation.

In the deterministic case described above, the reservation handshake alone is sufficient to prevent physically impossible states such as double occupation of a module. In circular production systems, however, decision logic is non-deterministic: learned policies, probabilistic models, and human operators acting under uncertainty all participate in shop floor decisions, and procedural guards cannot anticipate every failure mode they introduce. To demonstrate that the knowledge graph enforces physical consistency independently of application-logic correctness, the demonstrator includes a configurable fault-injection mode that deliberately bypasses the reservation protocol, allowing the SHACL constraint layer to be evaluated as an infrastructure-level safety net.

At runtime, each module instantiates its own semantic middleware as an embedded agent that handles service and workflow registration, REST API exposure, and knowledge-graph synchronization, while the module's control logic retains full authority over operational decisions. Peer modules discover each other's workflows through SPARQL queries via the interface layers triple store access module, retrieve the endpoint published in the service individual's address property, and invoke the workflow directly: no module holds a hard coded reference to any other. Every transfer handshake reads validated semantic state from and writes updates to the knowledge graph through the OGM, with the middleware routing the messages and the OGM determining their structure. Under normal operation, the reservation handshake ensures that a receiving module is unoccupied before conveyance proceeds, and the knowledge graph accepts the resulting possession update.

The two fault scenarios introduced earlier are realized as deliberate bypasses of this protocol. In the first scenario, a module skips the reservation check and initiates conveyance to a neighbor that already possesses a box. The OGM generates a SPARQL update that would add a second \texttt{:hasPossession} triple to the receiving module; GraphDB's SHACL engine evaluates \texttt{:FlexConveyorModuleShape}, detects that the \texttt{sh:maxCount~1} constraint on \texttt{:hasPossession} would be violated, and rejects the entire transaction before any triple is admitted. The validation report returned by GraphDB, shown in Listing~\ref{lst:shacl-violation}, identifies the focus node, the constraint path, the source shape, and the violation type. The graph state remains unchanged; the invalid triples are never persisted.

\begin{figure*}
\begin{lstlisting}[
  caption={SHACL-SPARQL shape \texttt{:InTransitBoxPossessedShape}, enforcing the cross-entity invariant that a box in the \texttt{:InTransit} state must be possessed by exactly one resource.},
  label={lst:shacl-intransit},
  basicstyle=\ttfamily\footnotesize,
  breaklines=true,
  frame=single
]
:InTransitBoxPossessedShape rdf:type sh:NodeShape ;
    sh:targetClass :Box ;
    sh:sparql [
        sh:message "A Box in InTransit state must be possessed
                    by exactly one resource." ;
        sh:prefixes [
            sh:declare [
                sh:prefix "fc" ;
                sh:namespace "http://w3id.org/circularfactory/FlexConveyor#"
                             ^^xsd:anyURI ;
            ]
        ] ;
        sh:select """
            SELECT $this
            WHERE {
                $this fc:hasState ?state .
                ?state rdf:type fc:InTransit .
                {
                    SELECT $this (COUNT(?possessor) AS ?count)
                    WHERE {
                        $this fc:isPossessedBy ?possessor .
                    }
                    GROUP BY $this
                }
                FILTER (?count != 1)
            }
        """ ;
    ] .
\end{lstlisting}
\end{figure*}

\begin{figure*}

\begin{lstlisting}[
  caption={SHACL validation report returned by GraphDB on illegal conveyance attempt, (abbreviated: Blank node hashes collapsed, URLs replaced by prefixes for readability) },
  label={lst:shacl-violation},
  basicstyle=\ttfamily\footnotesize,
  breaklines=true,
  frame=single
]
@prefix sh:       <http://www.w3.org/ns/shacl#> .
@prefix rdf:      <http://www.w3.org/1999/02/22-rdf-syntax-ns#> .
@prefix xsd:      <http://www.w3.org/2001/XMLSchema#> .
@prefix rdf4j:    <http://rdf4j.org/schema/rdf4j#> .
@prefix rdf4j-sh: <http://rdf4j.org/shacl-extensions#> .
@prefix fci:      <http://w3id.org/circularfactory/FlexConveyorInstances#> .
@prefix fc:       <http://w3id.org/circularfactory/FlexConveyor#> .

_:report  rdf:type        sh:ValidationReport ;
          sh:conforms     "false"^^xsd:boolean ;
          rdf4j:truncated "false"^^xsd:boolean ;
          sh:result       _:result .

_:result  rdf:type                     sh:ValidationResult ;
          sh:focusNode                 fci:Module3 ;
          sh:resultPath                fc:hasPossession ;
          sh:sourceConstraintComponent sh:MaxCountConstraintComponent ;
          sh:resultSeverity            sh:Violation ;
          sh:resultMessage             "A FlexConveyorModule may possess
                                        at most one Box at a time." ;
          rdf4j-sh:shapesGraph         rdf4j:SHACLShapeGraph ;
          sh:sourceShape               _:shape .

_:shape   rdf:type    sh:PropertyShape ;
          sh:path     fc:hasPossession ;
          sh:maxCount "1"^^xsd:integer ;
          sh:message  "A FlexConveyorModule may possess
                       at most one Box at a time." .
\end{lstlisting}
\end{figure*}

In the second scenario, a service updates a box's state to \texttt{:Delivered} without first removing its possession relation. The SPARQL-based shape \texttt{:DeliveredBoxNotPossessedShape} fires, detecting that a \texttt{:Delivered}-state box retains an \texttt{:isPossessedBy} triple. The transaction is rejected with a report identifying the cross-entity inconsistency: the box's lifecycle state and its possession relation are mutually incompatible under the formalized physical constraints. In both scenarios, the validation report propagates through GraphDB's REST API as a structured response, through the triple store access module as a typed Python exception, to the requesting service, providing a machine-readable explanation of which constraint was violated and why. The requesting asset receives the information necessary to adapt its behavior.

The enforcement mechanism positions the knowledge graph as an infrastructure-level safety net whose guarantees hold independently of application-logic correctness. SHACL shapes provide the authoritative, declarative constraint layer that no service writing to the knowledge graph can circumvent, even when application logic is prone to uncertainty.

This use case demonstrates more than SHACL-based constraint enforcement alone. It realizes a distributed production execution scenario in which autonomous modules register their capabilities in a shared knowledge graph, discover peer services through semantic queries, coordinate box transfers through event-driven workflows, and persist every state change as a validated update. The enforcement demonstrated here operates within a live execution context: SHACL evaluates the feasibility of state transitions across interrelated product and resource instances while distributed services actively read from and write to the knowledge graph, directly closing the gap identified in Section~\ref{sec:review}.
\FloatBarrier

\section{Evaluation}
\label{sec:evaluation}
\noindent 
Following Activity~5 of the DSRM~\cite{peffersDesignScienceResearch2007}, this section evaluates the KAPPS architecture against the 14 requirements of Section~\ref{sec:perspectives}, drawing on the use cases of Section~\ref{sec:demonstration} as observational evidence. Hevner et al.~\cite{hevnerDesignScienceInformation2004} treat evaluation as the matching of an artifact to a method appropriate to the claims being made. For KAPPS, the claim is architectural: the design of the system, not any single behavioral trace, is what satisfies the requirements. The evaluation is therefore organized around the architectural decisions from which satisfaction follows, with the use cases serving as observational witnesses. The evaluation rests on four architectural decisions:
\begin{enumerate}
\item The choice of an ontology governed RDF knowledge graph that extends the PPR framework, as the authoritative information state.
\item The Object-Graph Mapper as the single validated write path, through which every service commit reaches the graph as ontology-typed, OGM-validated data.
\item SHACL as the closed-world constraint layer operating server-side in the triple store, evaluating each admitted write against the physical invariants of the modeled domain.
\item The semantic middleware as the runtime layer carrying event-driven coordination, protocol mediation, and uniform service registration into the graph.
\end{enumerate}
The remainder of this section traverses the requirements in five clusters, each anchored on the decision or combination that satisfies it. Most requirements are carried by more than one decision in combination; R14, distinctively, is satisfied by a property of the substrate itself — its triple store-agnosticism — rather than by any single architectural layer.

Six requirements (R1-R6) are fulfilled structurally by one architectural decision: an RDF knowledge graph, informed by a set of aligned ontologies, is the authoritative information state for the entire shop floor, made operationally viable by the OGM, which projects ontology structure into validated data objects through which services interact with the graph. The decision answers the six requirements jointly: heterogeneous data is represented uniformly at the information level (R1) behind a single queryable interface (R2); persistent IRIs make every asset, operation, and observation uniquely referable (R4) and linkable to the type-level model (R6) and to the observations and provenance that determined its state (R5); and the same substrate hosts uncertainty representations as first-class typed attributes (R3). The OGM contributes the bi-directionality that makes this viable on the shop floor, providing the information-engineering analogue of the free-body diagram: a service receives the scoped subgraph its decision requires, operates on validated Python objects rather than SPARQL, and commits the result so that the decision is re-linked to the type, observation, and instance identity from which the subgraph was cut. Both use cases witness this pattern: UC1 across the perception, detection, and learning services sharing the screw-operation data, parameters and decision. UC2 across peer FlexConveyor modules sharing the reservation–handover state.

That re-integration discipline is what R7 and R9 express, and KAPPS enables SHACL validation to provide it: when a service commits its scoped subgraph back to the authoritative state, every update reaches the triple store through its SHACL engine. SHACL's closed-world semantics provide the rejection discipline OWL's open-world reasoning cannot — absence of a required precondition causes rejection, not deferral.
Two failure modes are distinguished. A plan can be infeasible against the current state (R7), so an operation is admitted whose preconditions are absent; or an individual state transition can violate a physical or procedural invariant at the moment it is written (R9), so execution produces a configuration the physical world does not admit. Both failure modes are handled by the same architectural mechanism.
UC2 witnesses this directly: every attempt to assign possession of more than one box to a module is caught by the triple store's SHACL engine, demonstrating that by providing the matching shapes graph, any infeasible state transition or planned operation can be prevented.

The deployment of an ontology-based triple store and the implementation of a bidirectional OGM ensure that KAPPS represents factory state uniformly and validates each commit against physical invariants. What they do not provide is an execution layer. A triple store holds state but does not push events; services reading and writing RDF are not inherently aware of one another's updates; industrial hardware speaks protocols the graph does not. These are MES-level concerns that R8, R10, and R11 reflect as requirements.
KAPPS addresses them through the semantic middleware runtime, which implements workflow orchestration, a connector framework, and a REST engine and binds them to OGM-derived data objects. The middleware handles service registry in the knowledge graph at startup; peers discover one another through SPARQL and invoke each other through REST, reading and writing the graph at semantically meaningful state transitions. Coordination over a shared validated view follows by construction (R10), and every commit can condition a peer workflow's next action, giving event-driven execution without a central orchestrator (R8). Protocol heterogeneity is absorbed separately, through a uniform four-method connector contract: protocol-specific messages are translated into OGM-materialized data objects at the connector boundary, so services above remain protocol-agnostic (R11).
Both use cases witness this. In UC2, three workflows form an event chain mediated through graph-state commits with no central orchestrator; in UC1, torque, axial-force, and position streams are ingested through the middleware's connector framework.

While the runtime execution layer described above determines how operations are \emph{executed}, R12 requires a specification of how they are \emph{formalized} as durable records. The PPR framework as formalized in~\cite{pfrommerOntologyRemanufacturingSystems2022} provides exactly that, and KAPPS supports it by design. By extending the operation class with provenance information tailored to the project's needs, KAPPS allows tracing back the actor, context, observed state, and outcome of every decision made during runtime operation (R12). UC1 demonstrates this by persisting an \texttt{:UnscrewingOperation} instance for every performed unscrewing, carrying \texttt{:hasResource} (actor), three \texttt{:TimeSeriesData} links (context), the active parameter snapshot (observed state), and \texttt{:hasSuccessStatus} (outcome) on a single first-class entity.
The same channel handles learned abstractions. Observations made during execution feed learning services that persist ontology-governed objects back into the knowledge base such as explicitly defined task models~\cite{dreherLearningSymbolicSubsymbolic2024} (R13) or, as demonstrated in UC1, updated learned parameters on type-level entities to reflect continuous improvement over time. Because these abstractions are written through the same OGM commit path as engineered values, downstream services do not need to distinguish between them.

R14 closes the temporal half of decision provenance on the graph the architecture described before has made authoritative. Because KAPPS interacts with that graph through SPARQL UPDATE alone, any triple store implementation that tracks changes at triple granularity satisfies R14; the architecture commits to none in particular. The demonstrator uses the GraphDB history plugin~\footnote{\url{https://graphdb.ontotext.com/documentation/11.3/data-history-and-versioning.html}}, which records every transactional change at triple granularity with commit timestamps and persists them across restarts. Past system states are therefore reconstructible through the same SPARQL interface used for live state.

Across the 14 requirements, KAPPS delivers a satisfied verdict on every one. The clustering shows what a requirement-by-requirement walk obscures: most are satisfied not by distinct mechanisms but by the four load-bearing architectural decisions acting jointly. No requirement is satisfied at the expense of another. Scope conditions under which these satisfactions hold are discussed in Section~\ref{sec:limitations}.

\section{Limitations and Outlook}
\label{sec:limitations}
\noindent 

The evaluation in Section~\ref{sec:evaluation} closes Activity 5 of the DSRM process for the present iteration of KAPPS: requirements R1-R14 fixed in Section~\ref{sec:perspectives} have been assessed against the demonstrations in Section~\ref{sec:demonstration}. Within its stated scope, the structural claim is closed; the validity of KAPPS as a CPPS architecture for the Circular Factory, defined by R1-R14, is therefore established at the scientific level. Industrial-grade completeness of the implementation is a separate matter and is not claimed here: KAPPS is functional, not complete. The remainder of this section delimits the scope within which the closure holds and names the iteration that follows from it.

The validity of that closure holds under four scope conditions that deserve to be made explicit. 

(1)~KAPPS assumes that every artifact exchanged between services (messages, control flows, state representations, learned abstractions) can be expressed as RDF. Anything that cannot be represented in the knowledge graph falls outside the architecture. However, this does not mean, that data, that is not persistable to a triple store, such as images and videos is not suited for processing within the KAPPS Architecture, but rather refers to the ability of describing the information this data contains in RDF.

(2)~Structural and intensional correctness of data reaching the graph is guaranteed, but data veracity in the physical sense is not: when a measurement service is declared in the ontology to deliver a length measurement, KAPPS ensures that every such measurement reaches the graph and is recognizable as a length measurement, but whether the reported value corresponds to physical reality remains, as with any industrial instrument, the responsibility of the measurement resource itself.

(3)~The Circular Factory is treated as the upper envelope of information-infrastructure complexity for manufactured-goods production known at the state of publication: every information-flow challenge present in a linear factory is considered to be present in a circular one by design, since a circular factory contains linear production from raw material as one of its operating modes.

(4) Concurrency and isolation guarantees on the write path are delegated to the triple store and are therefore properties of the chosen backend rather than of the architecture itself; identity, access control, and triple-level rights management are expected to be provided by the enclosing factory ecosystem rather than by KAPPS.

The step from scientific completeness to industrial deployment defines the next iteration of KAPPS and, under the DSRM process model, the Activity~5$\to$3 feedback from the evaluation of the present artifact into the design and development of its successor. In the short term, this iteration takes the form of an end-to-end deployment in a running micro circular factory for angle grinders, instrumented as a full-scale research prototype with physical transformer, inspection, and learning cells operating over a single knowledge graph. This deployment brings a problem set that the present paper does not address.

(1)~Domain experts must be equipped to author and maintain the ontologies and SHACL shapes the architecture consumes; this requires modeling workflows and tooling that lower the epistemic barrier from datasheet-and-checklist practice to ontology-and-shape practice, and is, in our current assessment, the single largest obstacle to industrial adoption. 
(2)~The knowledge graph must be embedded into an enclosing factory ecosystem that supplies identity, access control, and triple-level rights management; candidate systems include research data management platforms such as Kadi as well as established enterprise user-management stacks, any of which can provide the security surface KAPPS itself does not.

(3)~Ontology evolution during operation must be coordinated with the restart cycles that physical resources require when their functionality changes, so that schema updates propagate predictably at well-defined points rather than mid-execution.

(4)~Liveness under decentralized control must be supported by middleware-level heartbeat supervision, complemented by a triple store-side watchdog that removes stale discovery triples when a resource-bound middleware instance falls silent, so that planning services never commit to unreachable resources

(5)~The quantitative deployment envelope (write throughput under sustained load with server-side SHACL enforcement active, latency distributions across the OGM, scaling behavior of the triple store as the knowledge graph grows) must be characterized empirically under conditions an isolated demonstrator cannot produce.

The longer-term research horizon extends KAPPS along three directions, each building on the infrastructure-level position the architecture establishes.

(1)~The SHACL shapes that encode physical state-transition invariants are at present authored by engineers. Operator feedback captured during live execution provides a channel through which these shapes can be refined, extended, and ultimately learned, turning the constraint layer itself into a co-evolving artifact rather than a static specification.

(2)~The knowledge graph of a single factory becomes most valuable when it federates with the graphs of other factories, supply-chain partners, and regulatory bodies. Treating KAPPS deployments as nodes in a federated knowledge graph over Linked Open Data opens the architecture to cross-site traceability, multi-site planning, and circular-economy reporting without requiring central data consolidation. 

(3)~As autonomous AI agents may take on increasing portions of planning and decision-making inside manufacturing systems, the execution-time constraint enforcement KAPPS provides becomes a structural safeguard: the knowledge graph serves as auditable agent memory, the SHACL layer as a non-negotiable action boundary, and the ontology as the grounding through which agent reasoning remains connected to the physical state of the factory~\cite{klein_toward_2026}. These directions are not independent research programs; each presupposes the architectural substrate the present work establishes and evaluates, and each sharpens rather than replaces the claim closed here.

\color{black}

\section{Conclusion}
\label{sec:conclusion}
\noindent 
Conventional CPPS architectures, designed for static product structures and predefined sequences, do not meet the demands of circular production, in which heterogeneous core conditions, dynamically adaptable process flows, and the continuous integration of human and machine knowledge become the operational norm rather than the exception. This paper has addressed that gap along the four contributions stated in the introduction. The Circular Factory was analyzed through five complementary perspectives, yielding the 14 requirements R1-R14 that any supporting CPPS must satisfy. The KAPPS architecture was proposed as a four-layer, knowledge-graph-centric runtime architecture in which semantic models actively govern execution, validation, and state evolution and two demonstrations under representative Circular Factory conditions instantiated the architecture. The successful systematic evaluation against R1-R14 closed the DSRM cycle of the present iteration and the validity of KAPPS as a CPPS architecture for the Circular Factory is therefore established within the scope conditions delimited in Section~\ref{sec:limitations}. The core software artifacts and the demonstrator code for both use cases are released as open source alongside this paper, completing Activity~6 of the DSRM process and inviting the manufacturing-systems research community to reproduce, extend, and build upon the approach as it moves from functional completeness towards deployment in the next iteration.

\section*{CRediT authorship contribution statement}
\textbf{Etienne Hoffmann: } Conceptualization, Software, Validation, Visualization, Investigation, Methodology, Writing - Original Draft, Writing - Review \& Editing, Project Administration.
\textbf{Jan-Felix Klein: }Writing - Review \& Editing, Software, Visualization, Funding acquisition.
\textbf{Sören Weindel: }Writing - Review \& Editing, Software.
\textbf{Max Goebels: }Writing - Original Draft (sec.\ref{sec:uc1}), Writing - Review \& Editing, Software, Visualization.
\textbf{Sebastian Behrendt: }Writing - Review \& Editing.
\textbf{Daniel Hernández: }Writing - Review \& Editing.
\textbf{Ratan Bahadur Thapa: }Writing - Review \& Editing.
\textbf{Jürgen Fleischer: }Writing - Review \& Editing, Resources, Supervision, Funding acquisition.
\textbf{Kai Furmans: }Writing - Review \& Editing, Resources, Supervision, Funding acquisition.
\textbf{Steffen Staab: }Writing - Review \& Editing, Supervision, Funding acquisition.

\section*{Acknowledgements}
This work is funded by the German Research Foundation (DFG) - SFB 1574 – 471687386.

\section*{ Declaration of generative AI and AI-assisted technologies in the manuscript preparation process}
During the preparation of this work the authors used Claude in order to improve language and readability of the manuscript. After using this service, the authors reviewed and edited the content as needed and take full responsibility for the content of the published article.

\bibliographystyle{elsarticle-num} 
\bibliography{bibliography.bib}

\end{document}